%% file: main.tex
\documentclass[10pt,twocolumn,letterpaper]{article}

\usepackage[pagenumbers]{cvpr} % To force page numbers, e.g. for an arXiv version

\input{preamble}

\definecolor{cvprblue}{rgb}{0.21,0.49,0.74}
\usepackage[pagebackref,breaklinks,colorlinks,allcolors=cvprblue]{hyperref}

\title{Semantic Anchor Transport: Robust Test-Time Adaptation \\ for Vision-Language Models}

\author{Shambhavi Mishra$^*$ \quad Julio Silva-Rodríguez \quad Ismail Ben Ayed \\
Marco Pedersoli \quad Jose Dolz\\[1em]
LIVIA, ÉTS Montréal, Canada \\
International Laboratory on Learning Systems (ILLS), \\
McGILL - ETS - MILA - CNRS - Université Paris-Saclay - CentraleSupélec, Canada \\[0.5em]
$^*$ {\tt\small shambhavi.mishra.1@etsmtl.net}}
\begin{document}
\maketitle
\input{0_Abstract.tex}    
\input{1_Introduction.tex}
\input{2_Related.tex} 
\input{3_Method.tex} 
\input{4_Results.tex} 
\input{5_Conclusions.tex} 
{
    \small
    \bibliographystyle{ieeenat_fullname}
    \bibliography{main}
}

\input{6_Appendix.tex}
\end{document}

%% file: preamble.tex
\newcommand{\vx}{\mathbf{x}}

\usepackage{graphicx}	
\usepackage{amsmath}	
\usepackage{amssymb}	
\usepackage{booktabs}
\usepackage{times}
\usepackage{microtype}
\usepackage{epsfig}
\usepackage{caption}
\usepackage{float}
\usepackage{placeins}
\usepackage{color, colortbl}
\usepackage{stfloats}
\usepackage{enumitem}
\usepackage{tabularx}
\usepackage{xstring}
\usepackage{multirow}
\usepackage{xspace}
\usepackage{url}
\usepackage{amsmath}
\usepackage{natbib}
\usepackage[hang,flushmargin]{footmisc}
\usepackage{soul}
\usepackage{tikz}  % For drawing shapes
\usepackage{float}
\usepackage{wrapfig}

\definecolor{darkgreen}{RGB}{0,128,0}
\definecolor{darkred}{RGB}{175,0,0}
\definecolor{lightblue}{HTML}{87CEFA}
\definecolor{lightpurple}{HTML}{8470FF}
\definecolor{myblue}{RGB}{38, 145, 199} 
\definecolor{mygreen}{RGB}{38, 199, 149}

\newcommand{\R}[1]{{%
    \textbf{%
        \ifstrequal{#1}{1}{\textcolor{red}{R#1}}{%
        \ifstrequal{#1}{2}{\textcolor{blue}{R#1}}{%
        \ifstrequal{#1}{3}{\textcolor{magenta}{R#1}}{%
        \ifstrequal{#1}{4}{\textcolor{teal}{R#1}}{%
                           \textcolor{cyan}{R#1}%
        }}}}%
    }%
}}
\usepackage{xcolor}
\newcommand{\diff}[1]{\textcolor{green}{#1}}

\newcommand{\ours}{\textsc{SAT}}

\newcommand{\mypar}[1]{\noindent\textbf{#1}}

\newcommand{\appensecref}[1]{\textcolor{cvprblue}{Appendix~\ref{#1}}}

\newcommand{\Q}{\mathbf{Q}}

\newcommand{\Z}{\mathbf{Z}}
\newcommand{\T}{\mathbf{T}}
\newcommand{\vp}{\mathbf{p}}
\newcommand{\vq}{\mathbf{q}}
\newcommand{\vt}{\mathbf{t}}
\newcommand{\vu}{\mathbf{u}}
\newcommand{\vv}{\mathbf{v}}
\newcommand{\vz}{\mathbf{z}}

\newcommand{\Ent}{\mathcal{H}}
\newcommand{\Loss}{\mathcal{L}}
\newcommand{\real}{\mathbb{R}}

\usepackage{algorithm}
\usepackage{algpseudocode}

\usepackage[geometry]{ifsym}

\newcommand{\Ours}{SAT}   % Ours

\usepackage[table]{xcolor}  % for \cellcolor
\definecolor{GainBg}{RGB}{227,245,233}   % soft green background
\definecolor{BaseBg}{gray}{0.93}         % light gray for baseline cells
\definecolor{GainText}{RGB}{0,110,40}    % dark green text (readable in print)

\newcommand{\gain}[1]{\textbf{$\uparrow$\,#1}}

\newcommand{\gaintxt}[1]{{\textcolor{GainText}{\gain{#1}}}}

\usepackage{tikz}
\newcommand*\circled[1]{\tikz[baseline=(char.base)]{
            \node[shape=circle,draw,inner sep=1pt] (char) {#1};}}

%% file: 0_Abstract.tex
\begin{abstract}

Large pre-trained vision-language models (VLMs), such as CLIP, have shown unprecedented zero-shot performance across a wide range of tasks. Nevertheless, these models may be unreliable under distributional shifts, as their performance is significantly degraded. In this work, we investigate how to efficiently utilize class text information to mitigate distribution drifts encountered by VLMs during inference. In particular, we propose generating pseudo-labels for the noisy test-time samples by aligning visual embeddings with reliable, text-based semantic anchors. Specifically, to maintain the regular structure of the dataset properly, we formulate the problem as a batch-wise label assignment, which is efficiently solved using Optimal Transport. Our method, Semantic Anchor Transport (\ours{}), utilizes such pseudo-labels as supervisory signals for test-time adaptation, yielding a principled cross-modal alignment solution. Moreover, \ours{} further leverages heterogeneous textual clues, with a multi-template distillation approach that replicates multi-view contrastive learning strategies in unsupervised representation learning without incurring additional computational complexity. Extensive experiments on multiple popular test-time adaptation benchmarks presenting diverse complexity empirically show the superiority of \ours{}, achieving consistent performance gains over recent state-of-the-art methods, yet being computationally efficient.
\end{abstract}

%% file: 1_Introduction.tex
\section{Introduction}
\label{sec:intro}

Large pre-trained vision-language models (VLMs), such as CLIP \cite{radford2021learning} and ALIGN \cite{jia2021scaling}, have emerged as a new paradigm shift in machine learning, revealing promising zero-shot transferability. Nevertheless, if the model is exposed to domain drifts at test time, its performance can be largely degraded \cite{yu2023task,silva2024closer}. While a straightforward solution to bridge this gap involves fine-tuning the trained model using domain-specific labeled data \cite{lai2023padclip,goyal2023finetune}, this strategy presents several limitations in real-world scenarios, which may hinder its scalability and deployment. First, adapting to new domains requires collecting labeled samples drawn from each distinct distribution. This might be impractical for specific domains and further hinder the real-time adaptation of the trained model for each input test sample. Furthermore, fine-tuning the model may undermine its desirable zero-shot capabilities~\cite{WiSE}.

Test-Time Adaptation (TTA) presents a realistic and practical scenario for unsupervised domain adaptation, where a pre-trained model requires real-time adaptation to new data to address unknown distribution drifts without access to supervisory signals \cite{wangtent,iwasawa2021test,yuan2023robust}. Nevertheless, despite the recent rise of VLMs, and the popularity of TTA in more traditional deep models, such as CNNs and ViTs, the study of TTA in large pre-trained vision-language models remains less explored. Standard approaches, e.g., TENT \cite{wangtent}, which minimizes a Shannon entropy objective, have been adopted in CLIP test-time adaptation \cite{shu2022test}. On the other hand, more recent strategies utilize pseudo-labels within the inference batch to guide model adaptation \cite{osowiechi2024watt,hakim2024clipartt,maharana2024texttt}. In particular, the core idea of these later methods is to minimize a classification loss, typically the standard cross-entropy, between the generated pseudo-labels and model predictions, which guides the network updates over multiple iterations. Nonetheless, while this strategy is common in semi-supervised learning, where additional labeled instances are available for all categories, applying it naively in unsupervised scenarios, such as test-time adaptation, can lead to degenerate solutions over multiple updates \cite{iwasawa2021test}, a problem we refer to as \textit{error accumulation}.

\begin{figure}[h!]
    \centering
    \includegraphics[width=\columnwidth]{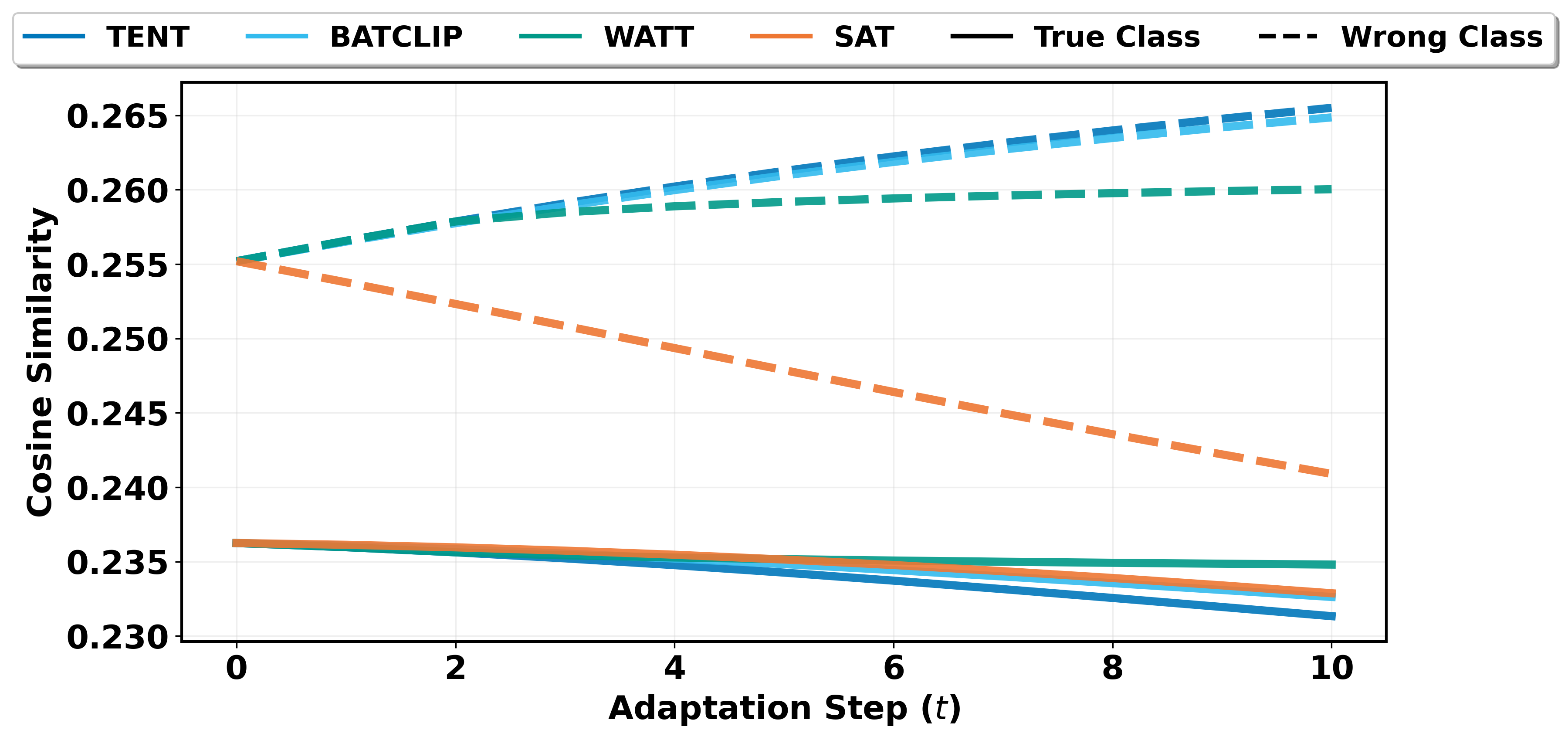}
    \caption{\textbf{Error Accumulation.} We track examples from all corruptions from CIFAR10C dataset with initial zero-shot CLIP predictions (at $t=0$) that are misclassified.
    %with confidence $> 0.7$. 
    Baselines \cite{wangtent, maharana2024texttt, osowiechi2024watt} catastrophically reinforce this error: similarity to the ‘Wrong Class’ (dashed lines) increases while similarity to the ‘True Class’ (solid lines) decreases. In contrast, our method, SAT (orange), is the only one that provides a corrective signal, actively reducing similarity to the wrong class and breaking the cycle of error accumulation.}
    \label{fig:teaser}
    %\vspace{-2.0em}
\end{figure}

This problem is particularly prevalent in the literature on test-time adaptation in VLMs. To illustrate this phenomenon, we conducted a simple experiment, the findings of which are depicted in Figure~\ref{fig:teaser}. Considering all corruptions in CIFAR10C, we identify samples misclassified by zero-shot CLIP and track their behavior across consecutive TTA steps. Specifically, we compute the average cosine similarity between the adapted visual embeddings and the corresponding class text embeddings of both the \textit{wrong} (i.e., predicted) and \textit{true} (i.e., expected) category. Ideally, a robust TTA method should reduce the similarity to the \textit{wrong} class prototype, gradually correcting the initial misalignment. Nevertheless, as shown in Figure~\ref{fig:teaser}, existing methods tend to reinforce these mistakes, thereby increasing the similarity between the visual and text prototypes of the incorrect class, even when the prediction is wrong.

In light of the above limitation, we aim to provide a more robust TTA framework that can correct incorrect initial supervisory signals. To do so, we reframe TTA as a cross-modal alignment problem, where the goal is \textit{to align the embeddings of test images to their corresponding fixed semantic anchors provided by the language modality}. This alignment can be viewed through the lens of clustering, where the unlabeled samples are grouped around class text prototypes. Hence, we can take into account the structure of the joint test batch to better correct sample-specific inconsistencies. Nevertheless, instead of jointly learning the cluster centroids during training, we use the fixed class prototypes derived from the text representations, which provide stable semantic guidance without requiring explicit supervision. A key challenge is the distribution mismatch between text prototypes and test image embeddings, since they originate from different modalities. Traditional clustering methods, which typically assume unimodal distributions, often struggle to bridge this gap, resulting in suboptimal performance. In contrast, we formulate label assignment as an Optimal Transport (OT) problem, yielding a global, cost-aware correspondence between visual embeddings and text prototypes that naturally model multi-modal distributions \cite{multimodal_dist_ot}. This perspective is more robust to outliers or noisy predictions, directly mitigating the error accumulation that undermines existing self-training in TTA.

\noindent The main contributions of this paper can be summarized as:

\begin{itemize}
    \item We propose Semantic Anchor Transport (SAT). This novel framework casts the pseudo-labeling strategy in CLIP test-time adaptation as a batch-wise Optimal Transport assignment, which leverages the class text information available in vision-language models in the form of fixed robust cluster centroids without requiring further annotations.

    \item To solve the label assignment task, we resort to the Sinkhorn algorithm, as it can handle multi-modal distributions and compute label assignments efficiently. 

    \item We introduce a multi-template knowledge distillation approach that leverages richer information derived from different text prompts to better guide adaptation without incurring significant computational overhead.

    \item Comprehensive experiments across multiple visual corruptions and domain shifts benchmarks demonstrate the superiority of \ours{} over recent state-of-the-art methods.
\end{itemize}

%% file: 2_Related.tex
\section{Related Work}

\mypar{Test-Time Adaptation (TTA)} aims at adapting a pre-trained model to a stream of incoming unlabeled target domain data, processed in batches during testing \cite{wangtent,zhang2022memo,choi2022improving,niutowards}. Existing approaches in traditional unimodal models can be roughly categorized into: \textit{i)} normalization-based methods, which leverage the statistics of the test data to adjust the BatchNorm statistics of the model \cite{mirza2022norm,schneider2020improving}; \textit{ii)} entropy-based approaches \cite{wangtent,niu2022efficient,goyal2022test}, where the model is adapted optimizing the Shannon entropy of the predictions; and \textit{iii)} pseudo-label strategies \cite{liang2020we}, which employ the test-time generated labels for supervising the model. With the advent of VLMs, several works have attempted to accommodate some of these techniques to adapt pre-trained foundation 
models, notably CLIP \cite{shu2022test,ma2024swapprompt,osowiechi2024watt,hakim2024clipartt}, which mainly differ on the different parameter groups updated during adaptation, i.e., prompt tuning \cite{shu2022test,ma2024swapprompt,dobler2024lost} or layer-norm \cite{hakim2024clipartt,osowiechi2024watt} strategies. For example, Test-Time Prompt Tuning (TPT) \cite{shu2022test} optimizes input text prompts by minimizing the model's prediction entropy. In contrast, Vision-Text-Space Ensemble (VTE)~\cite{dobler2024lost} uses an ensemble of prompts as input to the text encoder. Nevertheless, keeping both the text and vision encoder frozen makes it difficult for the model to adapt to images with severe noise effectively. Recent VLM adaptation methods, such as WATT~\cite{osowiechi2024watt} and BATCLIP~\cite{maharana2024texttt}, which employ parameter-efficient updates of the encoders, have developed sophisticated bimodal heuristics to improve pseudo-label quality. WATT \cite{osowiechi2024watt} generates them by combining both image-to-image and text-to-text similarity matrices. Taking this a step further, BATCLIP \cite{maharana2024texttt} first generates pseudo-labels via a standard argmax prediction, which are refined via additional losses. Nevertheless, both approaches rely on locally derived pseudo-labels, failing to explicitly optimize the sample-to-class assignment group cost, making them more sensitive to outliers or ambiguous samples. 

\mypar{Clustering for unsupervised representation learning.} Jointly adapting the parameters of a deep network while inferring the class assignments can be viewed as \textit{clustering} and \textit{unsupervised representation learning}. Thus, our work is closely related to recent literature on deep clustering-based approaches \cite{asanoself,caron2018deep,caron2019unsupervised,huang2019unsupervised,yan2020clusterfit,yang2016joint,jabi2019deep,caron2020unsupervised}. In \cite{caron2018deep}, $k$-means assignments are leveraged as pseudo-labels to learn visual representations, a strategy later employed in \cite{caron2019unsupervised} to pre-train standard supervised deep models. Nevertheless, applying naive $k$-means risks collapsing to only a few imbalanced clusters, making it ineffective. A more principled approach consists in framing the label assignment task as an instance of the Optimal Transport problem \cite{asanoself}, whose global, batch-aware optimization and balancing constraints naturally prevent degenerate solutions. Furthermore, Caron \textit{et al.} \cite{caron2020unsupervised} enforce consistency between cluster assignments across different image views or augmentations, avoiding expensive pairwise comparisons typically performed in contrastive learning. While our formulation shares similarities, key differences exist. In particular, all these methods, i.e., \cite{caron2019unsupervised,asanoself,caron2018deep} operate in unsupervised representation learning, where prototypes are learned from the data distribution. In contrast, we leverage the text representations as fixed semantic anchors. %to enhance the cluster assignments. 
Note that text embeddings are accessible \textit{for free} in VLMs, not incurring additional supervision, as the test image label remains unknown. A natural benefit from this strategy is that we do not need to resort to additional augmentation or ``multi-views" strategies \cite{caron2020unsupervised}, which introduce a computational burden. Instead, we treat individual text templates as ``augmented views", enhancing the representation learning of the adapted model without incurring extra cost. In fact, these embeddings are computed only once \textit{offline}, whereas %image augmentations require new forward passes for each batch at test time. %In contrast, 
image augmentation or ``multi-views" strategies must create the additional images for each incoming batch, performing one forward pass per new augmentation at test time. 

%% file: 3_Method.tex
\section{Background}

\subsection{Vision-language models}

CLIP \cite{radford2021learning} is a foundation vision-language model, trained via contrastive learning to produce visual representations from images $\vx$ paired with their associated text descriptions $T$. To do so, CLIP consists of an image encoder $\boldsymbol{\theta}$ and a text encoder $\boldsymbol{\phi}$. This generates the corresponding vision $\vz \in \real^d$ and class text $\vt_k \in \real^d$ embeddings (column vectors), which are typically projected into an $\ell_{2}$-normalized shared embedding space. At inference, this learning paradigm enables zero-shot prediction. More concretely, for a given set of $K$ classes, and an ensemble of $M$ different %prompts 
templates, %per category, 
we can generate the set of available prompts as $\mathcal{T}=\{\{T_{mk}\}_{m=1}^M\}_{k=1}^K$, whose embedding for template $m$ and class $k$ is obtained as $\vt_{mk}=\boldsymbol{\phi}(\textsc{\textcolor{gray}{``A photo of a [class$_k$]''}})$. Then, a popular strategy \cite{radford2021learning,clipad,WiSE} consists in obtaining a class zero-shot prototype, which is computed as $\vt_k=\frac{1}{M}\sum^M_{m=1}\boldsymbol{\phi}(T_{mk})$. Then, for a given test image $\vx_i$, its zero-shot prediction, $\vp_i=(p_{ik})_{1 \leq k \leq K}$, can be obtained as:
\begin{equation}
\label{eq:clip}
p(y=k|\vx_i)= %p_{ik} =
\frac{\exp{(\vz_i^\top \vt_k/\tau)}}{\sum_{j=1}^K\exp{(\vz_i^\top \vt_j/\tau)}},
\end{equation}

\noindent where $\tau$ is a temperature parameter %, whose value is 
learned during pre-training \cite{radford2021learning}, and $\vz^\top \vt$ indicates dot product. %operator.

\subsection{Test-Time Adaptation using CLIP}

\mypar{Problem setting.} We address the problem of adapting a pre-trained VLM at test time. In particular, given a model trained on the source domain $\mathcal{D}_S$, our goal is to adjust this model online to the new target domain $\mathcal{D}_T$, where only unlabeled test data $\{\vx_i\}^N_{i=1}$ is available, and which is received as a stream of batches, from which predictions must be provided.

\vspace{0.5mm}
\mypar{Vanilla entropy minimization.} Inspired by the semi-supervised learning literature, a straightforward solution to leverage the predicted probability in \cref{eq:clip} in TTA would be to adapt the model parameters based on an entropy minimization objective, similar to TENT~\cite{wangtent}:
\begin{equation}
\label{eq:tent}
    \Loss(p)=-\frac{1}{B_T}\sum_{i=1}^{B_T} \sum_{k=1}^K p(y=k|\vx_i) \log p(y=k|\vx_i),
\end{equation}

\noindent with $B_T$ denoting the size of the test batch. %, or batch. 
However, relying on the Shannon entropy \cite{wangtent} in the fully unsupervised case poses a significant risk, as it may potentially result in a degenerate solution, i.e., Eq. (\ref{eq:tent}) might be trivially minimized by assigning all data points to a single, arbitrary label.

\section{SAT: TTA as Cross-Modal Alignment}
\label{sec:method}

\begin{figure*}[t!]
    \centering
    \includegraphics[width=0.9\linewidth]{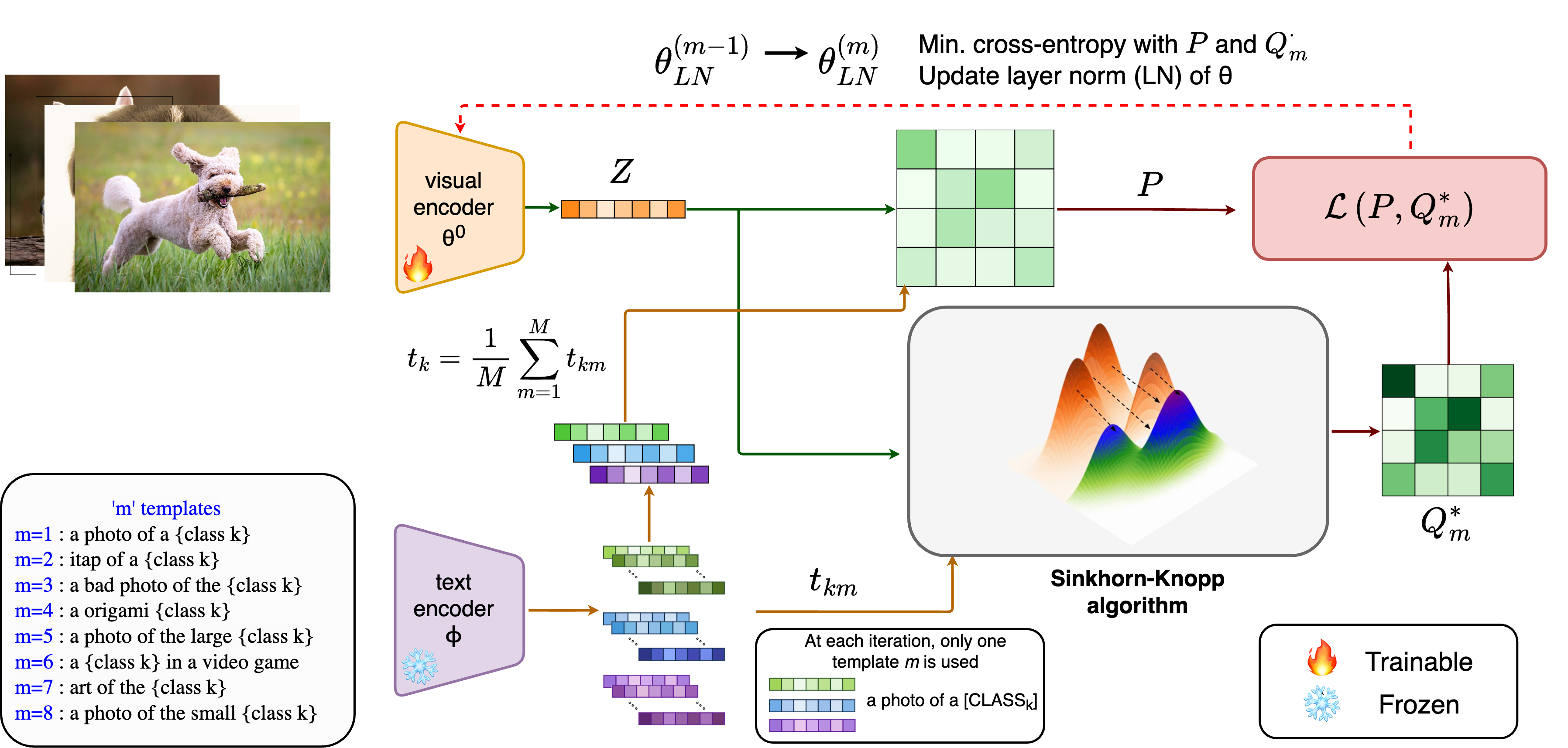}
    \caption{\textbf{\Ours{}} leverages Optimal Transport (with the Sinkhorn algorithm) to yield soft-codes $Q^*_m$ (\cref{eq:individual-codes}). Then, it minimizes the pseudo cross-entropy between $Q^*_m$ and the CLIP predictions $P$ as unsupervised loss during test-time adaptation (\cref{eq:posterior}). Note that, at each test batch, our model runs for $m$ iterations, each using a different text template $m$ to obtain the pseudo-codes $Q^*_m$.
    }
    \label{fig:method}
\end{figure*}

The proposed approach aims to overcome the limitations in \cref{eq:tent} objective by finding reliable pseudo-labels in a batch of test-time samples without supervision. To achieve this, we propose \textbf{\ours}. We formulate TTA as a cross-modal alignment problem, where the core task is to generate a robust supervisory signal for each unlabeled test batch. This novel TTA method leverages the text embeddings generated by the frozen CLIP text encoder more effectively, serving as strong class descriptors to yield label assignments for test samples, which we propose propagating through Optimal Transport. Note that our motivation drastically differs from the existing self-supervised representation learning literature that also resorts to pseudolabels \cite{caron2018deep}, where cluster centroids are iteratively updated based on the prior label assignments. Furthermore, by solving a globally optimal matching problem, \ours{} is more robust to noisy predictions than existing TTA methods \cite{osowiechi2024watt,maharana2024texttt}. Below, we elaborate on the details of our method, which is illustrated in Figure \ref{fig:method}.

To overcome the limitation of naively minimizing \cref{eq:tent} in the fully unlabeled scenario, TTA literature typically encodes the model predictions as posterior distributions $q(y|\vx_i)$, which results in:
\begin{equation}
\label{eq:posterior}
    \Loss(p,q)=-\frac{1}{B_T}\sum_{i=1}^{B_T} \sum_{k=1}^K q(y=k|\vx_i) \log p(y=k|\vx_i).
\end{equation}

Thus, the adaptation is driven by the pseudo cross-entropy in Eq. (\ref{eq:posterior}), where a significant challenge lies in constructing a high-quality target distribution, or pseudo-label $q_{ik}=q(y=k|\vx_i)$, without access to ground truth labels. Indeed, as illustrated in Figure~\ref{fig:teaser}, prior TTA methods tend to amplify prediction errors, leading to a state of \textit{error accumulation}. Our central thesis is that we can break this loop by exploiting the natural alignment between visual features and their text-based semantic anchors in the batch.

\subsection{Batch-Aware Cross-modal Alignment}
\label{sec:ot_solution}
% We address the problem at its source by abandoning the greedy, instance-wise prediction paradigm. Instead, we generate the entire pseudo-label matrix $\mathbf{Q} \in \mathbb{R}^{K \times B_s}$ for a batch in a single, globally optimal step. We formulate this as finding the optimal alignment between the batch's visual features $\mathbf{Z} \in \mathbb{R}^{d \times B_s}$ and a set of text-based \textbf{Semantic Anchors} $\mathbf{T} \in \mathbb{R}^{d \times K}$. 
In the following, we detail how to encode robust posteriors into \cref{eq:posterior} by exploiting multi-modal capabilities of CLIP and the test-time batch data distribution. Considering the probability is obtained as in \cref{eq:clip}, the objective in \cref{eq:posterior} can be expressed as:
%\noindent \textbf{Learning objective.} We propose to generate the cost assignment matrix $\mathbf{Q} \in \mathbb{R}^{K \times B_s}$ in a single, globally optimal step by solving a cross-modal matching problem. In particular, let us consider that the probability is obtained as in \cref{eq:clip}, the objective in \cref{eq:posterior} can be expressed as:
\begin{equation}
\label{eq:midstep}
\Loss(p,q)=-\frac{1}{B_T}\sum_{i=1}^{B_T}\left[\frac{1}{\tau}\vz_i^\top\T \vq_i-\log \sum_{k=1}^K \exp (\frac{\vz_i^{\top}\vt_k}{\tau})\right],
\end{equation}
\noindent where $\T=[\vt_1,...,\vt_K]$ represents the matrix containing the class text prototypes, and the logarithmic term in the right side does not depend on $\vq$. If we now express \cref{eq:midstep} in its matrix form over all the test images in the batch (with $\Z \in \mathbb{R}^{d \times B_T}$ the corresponding embeddings), and let $\Q$ be a matrix with columns $\vq_i$, we have the following objective:
\begin{equation}
\label{eq:ot_1}
    \max_{\Q \in \mathcal{Q}} \; tr(\Q^\top\T^\top\Z).
\end{equation}
\noindent which seeks the optimal alignment between the batch's visual features $\mathbf{Z} \in \mathbb{R}^{d \times B_T}$ and a set of text-based \textit{``semantic anchors"} $\mathbf{T} \in \mathbb{R}^{d \times K}$. Following \cite{cuturi2013sinkhorn}, we enforce the matrix $\Q$ to be an element of the \textit{transportation polytope}:
\begin{equation}
\label{eq:polytope-cons}
\mathcal{Q}\mathrel{:}= \{ \Q \in \real^{K \times B_T}_+ \mid \Q \mathbf{1}_{B_T} = \frac{1}{K} \mathbf{1}_K, \Q^\top \mathbf{1}_K = \frac{1}{B_T} \mathbf{1}_{B_T} \}, 
\end{equation}

\noindent with $\mathbf{1}_K$ and $\mathbf{1}_N$ denoting the vectors of ones in dimension $K$ and $N$, respectively. The constraints on $\mathcal{Q}$ enforce that on average each prototype is selected at least $\frac{N}{K}$ times, encouraging $\Q$ to be a matrix with uniform marginals in both rows and columns. \textit{Note that such constraints %directly produce dissimilarities 
‘break’ the direct
dependency between the original prediction $p$ in \cref{eq:tent} and the posterior $q$ in \cref{eq:midstep}, since the latter considers the whole batch distribution and its marginal properties for the assignment.} This results in more robust supervisory signals, later demonstrated empirically (Figure \ref{fig:ablation_components}).

% ‘Wrong Class’
%We frame this as finding the optimal alignment between the batch's visual features $\mathbf{Z} \in \mathbb{R}^{d \times B_s}$ and a set of text-based \textbf{Semantic Anchors} $\mathbf{T} \in \mathbb{R}^{d \times K}$. The objective is to find the assignment $\mathbf{Q}$ that maximizes the total similarity score:
%\begin{equation}
%\label{eq:ot_objective}
%\max_{\mathbf{Q} \in \mathcal{Q}} \; \text{tr}(\mathbf{Q}^\top\mathbf{T}^\top\mathbf{Z}),
%\end{equation}
%where $\mathcal{Q}$ is the \textit{transportation polytope} that constrains $\mathbf{Q}$ to represent a valid, balanced distribution.

\noindent \textbf{Finding optimal Q.} As the objective function in Eq. (\ref{eq:ot_1}) is linear, and the constraints defining $\mathcal{Q}$ are also linear, this is a linear program. Nevertheless, directly optimizing the above learning objective might be time-consuming, particularly as the number of data points and classes increases. To address this issue and facilitate faster optimization, we apply the Sinkhorn algorithm \cite{cuturi2013sinkhorn}, which incorporates an entropic constraint that enforces a simple structure on the optimal regularized transport. Hence, the optimization problem becomes:
\begin{equation}
\label{eq:main}
\max_{\Q \in \mathcal{Q}} \; tr(\Q^\top \T^\top \Z) + \varepsilon \Ent(\Q),
\end{equation}

\noindent where \( \Ent(\Q) \) denotes the Shannon entropy function $\Ent(\Q) = - \sum_{ij} \Q_{ij} \log \Q_{ij}$ and $\varepsilon$ controls its weight. Furthermore, to align with TTA assumptions, we work on batches by restricting the transportation polytope to the current batch \cite{caron2020unsupervised}, in contrast to other works that employ the full dataset \cite{asanoself}. Thus, the dimensionality of the $\Q$ matrix becomes $K \times B_T$, where $B_T$ denotes the batch size. Now, the soft assignment matrix $\Q^*$ is the solution of the problem %presented 
in (\ref{eq:main}) over the set $\mathcal{Q}$, which can be efficiently optimized with a few iterations: 
\begin{equation}
\label{eq:codes}
\Q^* = \text{Diag}(\vu^{(t)}) \exp\left( \frac{\T^\top \Z}{\lambda} \right) \text{Diag}(\vv^{(t)}),
\end{equation}
\noindent with $\vu$ and $\vv$ representing renormalization vectors in $\real^K$ and $\real^{B_T}$ respectively, and $t$ the iteration.

\noindent \textbf{Benefits of the proposed strategy.} The proposed Optimal Transport (OT) based formulation provides:
%\begin{enumerate}
    %\item 
    \textbf{i) Global consistency}, since OT is \textit{batch-aware}. The assignment for any single image is dependent on the features of all other images in the batch. The solver finds the best overall configuration, making it inherently robust to individual outliers. \textbf{ii) Intrinsic regularization} given by the polytope constraints (Eq. \ref{eq:polytope-cons}), which %$\mathcal{Q}\mathrel{:}= \{ \mathbf{Q} \in \mathbb{R}^{K \times B_T}_+ \mid \mathbf{Q} \mathbf{1}_{B_T} = \frac{1}{K} \mathbf{1}_K, \dots \}$, 
    enforce that, on average, all classes are represented in the pseudo-labels. This structurally prevents the model from collapsing to a single class, yielding robustness.
%\end{enumerate}

%To make this computationally feasible for online adaptation, we employ the standard entropic regularization strategy~\cite{cuturi2013sinkhorn}, allowing us to find the optimal transport plan $\mathbf{Q}^*$ with the highly efficient Sinkhorn-Knopp algorithm.

\subsection{Semantic alignment with multiple templates}
\label{sec:distillation}
% We leverage multiple text templates to generate diverse semantic cues, and distill them into a unified supervisory signal that guides robust label assignment at test time.
%The quality of the semantic anchors $\mathbf{T}$ is critical for better alignments. A single text prompt provides a narrow, potentially biased view of a class. To overcome this, TTA literature often resorts to a suboptimal approach, where the $M$ available text templates are simply averaged, which can blur the semantic signal. 
Recent literature \cite{osowiechi2024watt} has pointed out the limitations of using the average text prototypes, and has instead suggested to resort to multiple category embeddings, each obtained through the different text templates in $\mathcal{T}$. Thus, to create a richer and more robust supervisory signal, we leverage the ensemble of $M$ diverse text templates typically used in TTA, each providing a unique ``semantic view", $\mathbf{T}_m=[\vt_{1m},...,\vt_{Km}]$, where $\mathbf{T}_m$ denotes the matrix containing the class text prototypes obtained from the $m$-th template. %Indeed, TTA literature often resorts to a suboptimal approach, where the $M$ available text templates are simply averaged, which can blur the semantic signal. %A naive approach would be to simply average these views, but this can blur the semantic signal. 
Compared to existing TTA approaches, we introduce a more sophisticated knowledge distillation strategy that disentangles the goals of assignment and generalization.

\noindent \textbf{\circled{1} For Assignment, we need specificity.} We compute a separate, ``clean" transport plan $\mathbf{Q}^*_m$ for each specific semantic view $\mathbf{T}_m$. %=[\vt_{1m},...,\vt_{Km}]$, where $\mathbf{T}_m$ denotes the matrix containing the class text prototypes obtained from the $m$-th template. 
This provides an unambiguous matching based on a single, clear context (e.g., aligning to ``a sketch of a class" vs. ``a photo of a class"). Concretely, we modify the solution presented in \cref{eq:codes} to be specialized for specific class template\footnote{The text templates are the same as in WATT \cite{osowiechi2024watt} -- see \appensecref{appendix_dataset}.}, where the codes for each template $m$ are: 
\begin{equation}
\label{eq:individual-codes}
\Q^*_m = \text{Diag}(\vu^{(t)}_m) \exp\left( \frac{\T^\top_m \Z}{\lambda} \right) \text{Diag}(\vv^{(t)}_m).
\end{equation}

The renormalization vectors in our setting are computed trough a small number of matrix multiplications using the iterative Sinkhorn-Knopp algorithm \cite{knight2008sinkhorn}, where in each iteration $\vu^{(t)}_m=\vu_m/((\exp(\T_m^\top\Z/\lambda)\vv^{t-1}_m)$ and $\vv^{(t)}_m=\vv_m/((\exp(\T_m^\top\Z/\lambda)\vu^{t}_m)$, with $\vv^0_m=\mathbf{1}$. %Note that the computational burden introduced by OT is negligible, representing roughly $1\%$ of the total runtime for a batch, as discussed in Section \ref{ablation}.
    
\noindent \textbf{\circled{2} For Generalization, we need consistency.} To ensure generalization, model predictions must remain consistent across semantic views, preventing the visual encoder from overfitting to any single template, or ``view". %To enforce this i
In the proposed adaptation strategy, the model's prediction $p(y=k|\vx_i)$ in Eq. \ref{eq:clip} is always computed using a single %, consistent 
classification head, defined by the \textit{averaged} text class prototypes $\mathbf{T}$. %By fixing the prediction head to the averaged class prototypes, the predictions $p(y=k|\vx_i)$ remain stable across diverse templates. This 
This design enforces stability across templates, ensuring that the pseudo-supervision $q(y=k|\vx_i)$ is always applied against a consistent prediction function, preventing the encoder from overfitting to template‑specific fluctuations and promoting robust generalization.

%this consensus representation, the adaptation process is guided by stable semantics rather than %template‑specific fluctuations, ensuring that the cross‑entropy loss in Eq. (\ref{eq:posterior})} is minimized against a prediction function that does not vary across templates. 
    
\noindent \textbf{Coupling both together.} First, to optimize the objective in \cref{eq:posterior}, we employ the soft label assignments from (\ref{eq:individual-codes}), as they yield superior performance compared to their hard counterpart in other problems \cite{caron2020unsupervised}. Now, to distill the knowledge of the more diverse, and richer representations derived from multiple text predictions, the cross-entropy in \cref{eq:posterior} is optimized by iterating through each of the $M$ views, %each one using a different code $\boldsymbol{Q}_m$, 
which can be formally defined for each image $i$ and template $m$ as $\ell(\vp_i,\vq^*_{im})=-\vq^*_{im} \log \vp_i$, where $\vp_i$ is obtained with the average text template. Thus, for a given test image $i$ in a batch, after each update produced by the $m$-th text template, the visual encoder produces novel visual embeddings $\vz_i$, which are then used to generate new predictions $\vp_i$ for $i$. %In each step, a specific and clear supervisory signal ($\mathbf{Q}^*_m$) is used to guide the model towards learning a single, generalizable representation aligned with $\mathbf{T}$. This prevents the model from simply memorizing the quirks of individual prompts and instead forces it to learn the core, invariant features of a class.
%\end{itemize}
%During adaptation, we iterate through each of the $M$ views. In each step, a specific and clear supervisory signal ($\mathbf{Q}^*_m$) is used to guide the model towards learning a single, generalizable representation aligned with $\mathbf{T}$. This prevents the model from simply memorizing the quirks of individual prompts and instead forces it to learn the core, invariant features of a class.
%\vspace{-1mm}
%\subsection{The SAT Algorithm}
%\label{sec:algorithm}
%\noindent \textbf{The SAT Algorithm.} 

%\noindent \textbf{Whole learning strategy.} 
\subsection{Whole learning strategy}
The complete Semantic Anchor Transport (\ours) algorithm (Alg.~\ref{algo:ours}) consists in learning a label assignment matrix $\Q_m$ per text template by solving the optimization problem presented in Eq. (\ref{eq:main}), and then updating the affine parameters of the visual encoder's Layer Normalization layers, such as \cite{wangtent,osowiechi2024watt,hakim2024clipartt}. This is done iteratively by alternating two steps across the batches and text templates: \textit{i)} with the learnable parameters fixed, we compute the visual features $\mathbf{Z} \in \real^{d \times B_s}$ of the test batch samples, and find $\Q_m^*$ through Eq. (\ref{eq:individual-codes}) by iterating the updates on $\vu_m$ and $\vv_m$; and \textit{ii)} given %the current 
label assignments $\Q_m^*$, and the softmax predictions for the test samples obtained with each average class prototype $\vt_k$, the set of learnable parameters is updated by minimizing Eq. (\ref{eq:posterior}) w.r.t. the layer norm parameters, where stochastic gradient descent is applied over the whole batch. This process is repeated $M$ times. Finally, the class predictions on the batch are inferred with the updated model. %A view of t
%The proposed method is detailed in Alg.~\ref{algo:ours}.%(\appensecref{sec:clipot_alg}).

%The complete Semantic Anchor Transport (SAT) algorithm, which adapts the affine parameters of the Layer Normalization layers, as %For each test batch, we perform $M$ updates, one for each semantic view, as 
%detailed in Algorithm~\ref{algo:ours}. 

%\jose{This process effectively grounds the visual adaptation in a stable and diverse set of semantic anchors, providing a principled mechanism to mitigate error accumulation (Fig.~\ref{fig:teaser}) and deliver robust performance under real-world domain shifts. The resulting OT assignment matrix can be interpreted as a form of soft pseudo-labeling. Yet, unlike traditional self-training, it is derived from a globally optimal alignment (rather than sample-wise predictions), mitigating local error propagation typical of discrete pseudo-label updates.}

\begin{algorithm}[htbp]
\caption{Semantic Anchor Transport for one test batch.}
\label{algo:ours}
\begin{algorithmic}[1]
\State \textbf{Input:} Test batch $\{\vx_i\}_{i=1}^{B_T}$, set of $M$ semantic anchor matrices $\{\mathbf{T}_m\}_{m=1}^M$, visual encoder $\boldsymbol{\theta}$.
\State \textcolor{gray}{// $\mathbf{T}_m$ (and thus $\mathbf{T}$) are pre-computed offline from a systematically generated set of text templates.}
%\Statex
\State $\mathbf{T} \leftarrow \frac{1}{M}\sum_{m=1}^M \mathbf{T}_m$
\State \textcolor{gray}{\textbf{// --- Adaptation Phase ---}}
\For{each template $m$ in a random permutation of $\{1, \dots, M\}$}
    \State \textcolor{gray}{// Step 1: Align - Compute soft assignments that maximize cross-modal alignment for the m-\textit{th} ``view"}
    \State $\mathbf{Z} \leftarrow [\boldsymbol{\theta}(\vx_1), \dots, \boldsymbol{\theta}(\vx_{B_T})]$
    \State $\mathbf{Q}^*_m \leftarrow \text{SolveOT}(\mathbf{Z}, \mathbf{T}_m) \quad$ \textcolor{blue}{Eq. \ref{eq:individual-codes}} %\textcolor{blue}{Eq. \ref{eq:individual-codes}
    %\Statex
    \State \textcolor{gray}{// Step 2: Adapt - Distill knowledge into a general representation}

    \State $\mathbf{P} \leftarrow \text{Predict}(\mathbf{Z}, \mathbf{T}) \quad$ \textcolor{blue}{Eq. \ref{eq:clip}} 
    \State $\mathcal{L} \leftarrow \text{CrossEntropy}(\mathbf{P}, \mathbf{Q}^*_m)$
    \State Update LayerNorm parameters of $\boldsymbol{\theta}$ using $\nabla_{\boldsymbol{\theta}} \mathcal{L}$.
\EndFor
%\Statex
\State \textcolor{gray}{\textbf{// --- Inference Phase ---}}
\State $\mathbf{Z}_{\text{adapted}} \leftarrow [\boldsymbol{\theta}_{\text{adapted}}(\vx_1), \dots, \boldsymbol{\theta}_{\text{adapted}}(\vx_{B_s})]$
\State $\mathbf{P}_{\text{final}} \leftarrow \text{Predict}(\mathbf{Z}_{\text{adapted}}, \mathbf{T}) \quad$ \textcolor{blue}{Eq. \ref{eq:clip}}
\State \textbf{Return} $\mathbf{P}_{\text{final}}$
\end{algorithmic}
\end{algorithm}

%% file: 4_Results.tex
\section{Experiments}

\subsection{Setting}
\label{ssec:setting}

\mypar{Datasets.} 
We evaluate on standard TTA benchmarks \cite{maharana2024texttt, osowiechi2024watt, hakim2024clipartt}, covering two primary types of domain shift: \textbf{\textit{i})~Visual corruptions}: CIFAR-10C, CIFAR-100C, ImageNet-C, and Tiny-ImageNet-C \cite{datasets_corruptions}; and \textbf{\textit{ii})~Domain generalization}: PACS \cite{pacs}, VLCS \cite{vlcs}, OfficeHome \cite{officehome}, and VisDA-C \cite{visda}. We also use the original clean benchmarks (e.g., CIFAR-10 \cite{cifar}, CIFAR-10.1 \cite{cifar101}, and Tiny-ImageNet \cite{TinyImagenet}) for ablation. Full dataset details are presented in \appensecref{appendix_dataset}.

\mypar{Baselines:} 
We resort to relevant CLIP-based TTA parameter-efficient adaptation approaches, i.e., modify a set of learnable parameters, including TENT \cite{wangtent}, SAR \cite{niutowards}, VTE \cite{dobler2024lost}, TPT \cite{shu2022test}, CLIPArTT \cite{hakim2024clipartt}, WATT \cite{osowiechi2024watt}, and BATCLIP \cite{maharana2024texttt}, where \cite{hakim2024clipartt,osowiechi2024watt, maharana2024texttt} represent the state-of-the-art.

\mypar{Architectures:} Following prior work, we use Vision Transformer (ViT) backbones from CLIP \cite{radford2021learning}. Our main experiments use ViT-B/32 as in \cite{hakim2024clipartt,osowiechi2024watt}, and we demonstrate generalization with additional evaluations on larger ViT-B/16 and L/14 backbones, particularly CLIP's and SigLIP's \cite{zhai2023sigmoidlosslanguageimage}.

\mypar{Implementation details:} 
Following prior work \cite{hakim2024clipartt, osowiechi2024watt}, we use a batch size of 128 for all experiments (see \appensecref{sec:app_batch_size} for a sensitivity analysis on batch size). All reported results are the average of three runs using different random seeds.
We follow the officially recommended hyperparameters for all baseline methods. Specifically, TENT \cite{wangtent} and CLIPArTT \cite{hakim2024clipartt} are run for 10 adaptation steps per batch. For WATT \cite{osowiechi2024watt}, we use their suggested $L=2$, $M=5$, and 8 templates, totaling 80 iterations per batch. For BATCLIP \cite{maharana2024texttt}, we adapt both the text and visual encoders. For all other baselines, we use the learning rates specified in their respective papers for each dataset.
We set the learning rate for our method to $10^{-4}$ for all datasets, except for ImageNet-C, where we use $5 \times 10^{-5}$ as in \cite{maharana2024texttt}. Our formulation introduces only two new hyperparameters: the entropic constraint weight, $\epsilon$ from \cref{eq:main}, and the number of Sinkhorn iterations, $T$. These are fixed across all datasets and experiments. We set $\epsilon = 0.7$ (see \appensecref{epsilon_study}) and $T = 3$. We found $T=3$ to be sufficient for convergence, as in prior work \cite{caron2018deep}.

\subsection{Main results}

\mypar{Performance under common visual corruptions.} We present a unified evaluation across four standard benchmarks in Table \ref{tab:unified_main}, which shows the superiority of \Ours{} in adapting CLIP in the presence of common corruptions against a comprehensive suite of recent TTA methods. Compared to vanilla CLIP, our model brings performance gains of $17.8\%$ (CIFAR-10C) and $17.9\%$ (CIFAR-100C) without requiring additional supervision. These performance gains are similar when compared to other popular TTA methods, e.g., TENT or TPT, with differences ranging from $10\%$ to $21\%$. While this gap is reduced compared to recent approaches, such as WATT, CLIPArTT, and BATCLIP, the differences remain significant. \Ours{} outperforms the second-best competitor, i.e., WATT, by up to $3.2\%$ in CIFAR-10C (e.g., $5.1\%$ in \textit{pixelate}) and $1.8\%$ in the more challenging scenario of CIFAR-100C (which contains $\times$10 classes). Additionally, it achieves significant gains over recent baselines in specific corruptions, e.g., $19.8\%$ compared to BATCLIP in \textit{pixelate} (CIFAR-10C) or $16.3\%$ in \textit{Glass Blur} (CIFAR-100C). Also, compared to CLIPArTT, \Ours{} shows gains of nearly $10.6\%$ on \textit{pixelate} (CIFAR-100C) and $8.1\%$ on \textit{Gaussian Noise} (CIFAR-100C).

\begin{table*}[t]
\setlength{\tabcolsep}{4pt} % Adjust column spacing (was 3pt)
\centering
\scriptsize % Use scriptsize (was footnotesize)
\caption{\textbf{Performance in visual corruptions benchmarks.} Results using CLIP ViT-B/32. 
Best method in \textbf{bold}, second best \underline{underlined}.}
\vspace{-1em}
\label{tab:unified_main}
\begin{tabular}{
llccccccccccccccc|c
}
\toprule
\textbf{Method} & &
\rotatebox{90}{Gaussian} &
\rotatebox{90}{Shot} &
\rotatebox{90}{Impulse} &
\rotatebox{90}{Defocus} &
\rotatebox{90}{Glass} &
\rotatebox{90}{Motion} &
\rotatebox{90}{Zoom} &
\rotatebox{90}{Snow} &
\rotatebox{90}{Frost} &
\rotatebox{90}{Fog} &
\rotatebox{90}{Bright} &
\rotatebox{90}{Contrast} &
\rotatebox{90}{Elastic} &
\rotatebox{90}{Pixelate} &
\rotatebox{90}{JPEG} &
\textbf{Mean $\uparrow$} \\
\midrule
\multicolumn{17}{l}{\textbf{CIFAR-10C}} \\
\cmidrule(r){1-18}
CLIP & \textcolor{gray}{[ICLR'21]} & 35.27 & 39.67 & 42.61 & 69.76 & 42.40 & 63.97 & 69.83 & 71.78 & 72.86 & 67.04 & 81.87 & 64.37 & 60.83 & 50.53 & 55.48 & 59.22 \\
TENT & \textcolor{gray}{[ICLR'21]} & 41.27 & 47.20 & 48.58 & 77.12 & 52.65 & 71.25 & 76.20 & 78.29 & 79.84 & 77.39 & \underline{87.78} & 79.47 & 70.00 & 63.74 & 62.64 & 67.56 \\
SAR & \textcolor{gray}{[ICLR'22]} & 47.58 & 50.39 & 47.19 & 71.65 & 49.34 & 70.27 & 72.63 & 71.66 & 72.82 & 69.48 & 82.34 & 70.54 & 60.98 & 48.07 & 58.48 & 62.89 \\
VTE & \textcolor{gray}{[ECCVw'24]} & 44.40 & 47.70 & 42.90 & 64.90 & 45.00 & 66.70 & 67.00 & 67.40 & 64.50 & 65.30 & 74.90 & 53.60 & 61.20 & 42.60 & 50.80 & 57.26 \\
TPT & \textcolor{gray}{[NeurIPS'22]} & 33.90 & 38.20 & 37.66 & 67.83 & 38.81 & 63.39 & 68.95 & 70.16 & 72.39 & 64.31 & 81.30 & 62.26 & 56.43 & 42.80 & 53.67 & 56.80 \\
WATT & \textcolor{gray}{[NeurIPS'24]} & \underline{63.84} & \underline{65.28} & \underline{58.64} & \underline{78.94} & \underline{65.12} & \underline{77.81} & \underline{79.32} & \underline{79.79} & \underline{80.54} & \underline{78.53} & 87.11 & \underline{81.20} & \underline{72.66} & \underline{71.11} & \underline{67.36} & \underline{73.82} \\
CLIPArTT & \textcolor{gray}{[WACV'25]} & 59.90 & 62.77 & 56.02 & 76.74 & 61.77 & 76.01 & 77.40 & 77.29 & 79.20 & 75.74 & 86.59 & 77.82 & 70.20 & 66.52 & 63.51 & 71.17 \\
BATCLIP & \textcolor{gray}{[ICCV'25]} & 50.89 & 56.01 & 54.35 & 76.17 & 56.11 & 74.71 & 76.01 & 77.74 & 79.33 & 75.87 & 86.46 & 78.65 & 68.76 & 56.41 & 61.79 & 68.62 \\
\cellcolor{gray!20} \textbf{\Ours{} (\textit{Ours})} &
\cellcolor{gray!20} &
\cellcolor{gray!20}\textbf{64.85} & \cellcolor{gray!20}\textbf{67.34} & \cellcolor{gray!20}\textbf{62.27} & \cellcolor{gray!20}\textbf{82.09} & \cellcolor{gray!20}\textbf{68.07} & \cellcolor{gray!20}\textbf{81.30} & \cellcolor{gray!20}\textbf{83.13} & \cellcolor{gray!20}\textbf{83.71} & \cellcolor{gray!20}\textbf{83.40} & \cellcolor{gray!20}\textbf{82.56} & \cellcolor{gray!20}\textbf{89.90} & \cellcolor{gray!20}\textbf{84.86} & \cellcolor{gray!20}\textbf{76.08} & \cellcolor{gray!20}\textbf{76.25} & \cellcolor{gray!20}\textbf{70.03} & \cellcolor{gray!20}\textbf{77.06} \\
\midrule
\multicolumn{17}{l}{\textbf{CIFAR-100C}} \\
\cmidrule(r){1-18}
CLIP & \textcolor{gray}{[ICLR'21]} & 14.80 & 16.03 & 13.85 & 36.74 & 14.19 & 36.14 & 40.24 & 38.95 & 40.56 & 38.00 & 48.18 & 29.53 & 26.33 & 21.98 & 25.91 & 29.43 \\
TENT & \textcolor{gray}{[ICLR'21]} & 14.38 & 17.34 & 10.03 & 49.05 & 3.71 & 46.62 & 51.84 & 46.71 & 44.90 & 47.31 & 60.58 & 45.90 & 33.09 & 26.47 & 29.89 & 35.19 \\
SAR & \textcolor{gray}{[ICLR'22]} & 22.82 & 25.10 & 18.68 & 44.51 & 21.78 & 43.04 & 47.04 & 46.75 & 47.34 & 44.62 & 57.00 & 42.17 & 31.51 & 25.09 & 30.83 & 36.55 \\
VTE & \textcolor{gray}{[ECCVw'24]} & 10.00 & 10.30 & 13.30 & 36.10 & 20.40 & 37.90 & 39.80 & 42.20 & 40.80 & 36.60 & 45.50 & 29.20 & 30.80 & 17.00 & 20.70 & 28.71 \\
TPT & \textcolor{gray}{[NeurIPS'22]} & 14.03 & 15.25 & 13.01 & 37.60 & 16.41 & 37.52 & 42.99 & 42.35 & 43.31 & 38.81 & 50.23 & 28.09 & 28.12 & 20.43 & 28.82 & 30.46 \\
WATT & \textcolor{gray}{[NeurIPS'24]} & \underline{32.07} & \underline{34.36} & \underline{30.33} & \underline{52.99} & \underline{32.15} & \underline{50.53} & \underline{55.30} & \underline{52.77} & \underline{53.79} & \underline{51.49} & \underline{63.57} & \underline{52.76} & \underline{40.90} & \underline{40.97} & \underline{39.59} & \underline{45.57} \\
CLIPArTT & \textcolor{gray}{[WACV'25]} & 25.32 & 27.90 & 25.62 & 49.88 & 27.89 & 47.93 & 52.70 & 49.72 & 49.63 & 48.77 & 61.27 & 48.55 & 37.45 & 33.88 & 36.07 & 41.51 \\
BATCLIP & \textcolor{gray}{[ICCV'25]} & 17.25 & 19.76 & 18.98 & 42.20 & 19.00 & 40.81 & 46.59 & 41.34 & 40.14 & 41.56 & 53.85 & 34.07 & 31.38 & 25.51 & 28.96 & 33.43 \\
\cellcolor{gray!20} \textbf{\ours{} (\textit{Ours})} & 
\cellcolor{gray!20} &
\cellcolor{gray!20}\textbf{33.43} & \cellcolor{gray!20}\textbf{35.60} & \cellcolor{gray!20}\textbf{30.94} & \cellcolor{gray!20}\textbf{53.87} & \cellcolor{gray!20}\textbf{35.26} & \cellcolor{gray!20}\textbf{52.77} & \cellcolor{gray!20}\textbf{56.71} & \cellcolor{gray!20}\textbf{54.30} & \cellcolor{gray!20}\textbf{54.92} & \cellcolor{gray!20}\textbf{53.57} & \cellcolor{gray!20}\textbf{64.43} & \cellcolor{gray!20}\textbf{55.01} & \cellcolor{gray!20}\textbf{43.79} & \cellcolor{gray!20}\textbf{44.51} & \cellcolor{gray!20}\textbf{40.83} & \cellcolor{gray!20}\textbf{47.33} \\
\midrule
\multicolumn{17}{l}{\textbf{Tiny-ImageNet-C}} \\
\cmidrule(r){1-18}
CLIP & \textcolor{gray}{[ICLR'21]} & 7.08 & 9.41 & 3.44 & 21.71 & 9.12 & 34.52 & 27.44 & 32.51 & 36.33 & 25.94 & 40.15 & 1.81 & 30.40 & 22.78 & 29.59 & 22.15 \\
TENT & \textcolor{gray}{[ICLR'21]} & 8.01 & 10.04 & 4.18 & 24.53 & 10.09 & 36.94 & 29.48 & 32.20 & 35.72 & 27.46 & 39.79 & 2.24 & 31.92 & 24.79 & 30.93 & 23.22 \\
SAR & \textcolor{gray}{[ICLR'22]} & 9.09 & 10.94 & 3.65 & 5.50 & 1.68 & 14.02 & 12.08 & 20.72 & 24.62 & 8.37 & 32.35 & 0.71 & 15.32 & 12.39 & 25.35 & 13.12 \\
VTE & \textcolor{gray}{[ECCVw'24]} & \underline{18.63} & \underline{20.34} & 4.71 & 9.62 & 2.21 & 30.37 & 21.68 & 38.84 & 40.27 & 17.41 & 41.22 & 0.63 & 31.64 & 25.33 & 37.79 & 22.71 \\
TPT & \textcolor{gray}{[NeurIPS'22]} & 9.29 & 11.70 & 4.85 & 27.56 & 11.03 & 38.97 & 34.29 & 34.45 & 37.13 & 28.89 & 43.31 & 3.15 & 33.88 & 27.70 & 33.60 & 25.32 \\
WATT & \textcolor{gray}{[NeurIPS'24]} & 13.02 & 15.94 & 6.90 & 29.91 & 14.01 & 41.26 & 33.96 & 37.76 & \underline{39.65} & 32.13 & 46.93 & 3.53 & \underline{35.01} & 31.55 & 36.46 & 27.87 \\
CLIPArTT & \textcolor{gray}{[WACV'25]} & 14.44 & 17.44 & 10.37 & 31.46 & \underline{15.84} & 41.34 & 35.06 & 36.86 & 38.20 & \underline{33.44} & 46.43 & \underline{6.24} & 33.89 & 34.85 & 37.32 & 28.88 \\
BATCLIP & \textcolor{gray}{[ICCV'25]} & 11.96 & 15.48 & \underline{10.05} & \underline{31.89} & 14.76 & \underline{43.31} & \underline{39.07} & \underline{39.02} & 39.05 & 31.91 & \underline{49.06} & 5.65 & 32.79 & \underline{36.63} & \underline{39.12} & \underline{29.32} \\
\cellcolor{gray!20} \textbf{\Ours{} (\textit{Ours})} & \cellcolor{gray!20} &
\cellcolor{gray!20}\textbf{21.40} & \cellcolor{gray!20}\textbf{24.90} & \cellcolor{gray!20}\textbf{17.34} & \cellcolor{gray!20}\textbf{35.39} & \cellcolor{gray!20}\textbf{21.16} & \cellcolor{gray!20}\textbf{46.26} & \cellcolor{gray!20}\textbf{40.93} & \cellcolor{gray!20}\textbf{42.32} & \cellcolor{gray!20}\textbf{44.97} & \cellcolor{gray!20}\textbf{38.60} & \cellcolor{gray!20}\textbf{53.10} & \cellcolor{gray!20}\textbf{11.88} & \cellcolor{gray!20}\textbf{40.73} & \cellcolor{gray!20}\textbf{41.84} & \cellcolor{gray!20}\textbf{42.86} & \cellcolor{gray!20}\textbf{34.91} \\
\midrule
\multicolumn{17}{l}{\textbf{ImageNet-C}} \\
\cmidrule(r){1-18}
CLIP & \textcolor{gray}{[ICLR'21]} & 11.30 & 11.58 & 12.28 & 20.88 & 8.92 & 19.78 & 17.62 & 19.92 & 23.48 & 25.90 & 47.34 & 15.48 & 17.02 & 28.00 & 27.60 & 20.47 \\
TENT & \textcolor{gray}{[ICLR'21]} & 8.00 & 7.20 & 9.20 & 23.04 & 10.84 & 22.86 & 19.04 & 21.24 & 23.86 & 26.54 & 48.54 & 18.32 & 17.90 & 30.32 & 29.66 & 21.10 \\
SAR & \textcolor{gray}{[ICLR'22]} & 13.07 & \underline{15.69} & 13.92 & 22.74 & 14.53 & 23.41 & 19.49 & 22.65 & 24.89 & 29.47 & 48.39 & \underline{18.88} & 19.61 & 31.68 & 29.07 & 23.17 \\
VTE & \textcolor{gray}{[ECCVw'24]} & 7.12 & 10.24 & 9.18 & \textbf{27.31} & 10.27 & \textbf{26.42} & \textbf{27.36} & 24.28 & 26.15 & 31.22 & 49.37 & 13.09 & 14.18 & 32.44 & 31.33 & 22.66 \\
TPT & \textcolor{gray}{[NeurIPS'22]} & 8.94 & 7.22 & 7.55 & 20.47 & 9.13 & 21.78 & 23.92 & 24.61 & 21.54 & 24.98 & 40.37 & 15.22 & 13.18 & 30.74 & 24.63 & 20.01 \\
WATT & \textcolor{gray}{[NeurIPS'24]} & 7.76 & 7.06 & 8.94 & 24.16 & 12.46 & 25.00 & 21.52 & 21.58 & 24.16 & 26.62 & \textbf{49.74} & \textbf{21.14} & 19.90 & \underline{32.70} & \underline{32.16} & 22.33 \\
CLIPArTT & \textcolor{gray}{[WACV'25]} & 14.74 & 15.10 & 15.30 & 10.82 & 9.02 & 13.82 & 12.30 & 16.96 & 22.52 & 19.90 & 41.78 & 0.26 & 12.84 & 22.80 & 31.94 & 18.15 \\
BATCLIP & \textcolor{gray}{[ICCV'25]} & \underline{14.84} & 15.10 & \underline{15.52} & 24.42 & \underline{17.18} & 25.64 & 23.08 & \underline{25.06} & \underline{25.58} & \textbf{31.08} & \underline{49.66} & 18.44 & \underline{22.20} & \textbf{33.42} & \textbf{33.02} & \underline{24.95} \\
\cellcolor{gray!20} \textbf{\Ours{} (\textit{Ours})} & \cellcolor{gray!20} & \cellcolor{gray!20}\textbf{18.98} & \cellcolor{gray!20}\textbf{24.54} & \cellcolor{gray!20}\textbf{25.04} & \cellcolor{gray!20}\underline{25.46} & \cellcolor{gray!20}\textbf{21.98} & \cellcolor{gray!20}\underline{26.34} & \cellcolor{gray!20}\underline{24.46} & \cellcolor{gray!20}\textbf{27.78} & \cellcolor{gray!20}\textbf{28.74} & \cellcolor{gray!20}\underline{29.32} & \cellcolor{gray!20}41.98 & \cellcolor{gray!20}17.90 & \cellcolor{gray!20}\textbf{25.70} & \cellcolor{gray!20}26.32 & \cellcolor{gray!20}25.68 & \cellcolor{gray!20}\textbf{26.01} \\
\bottomrule
\end{tabular}
\end{table*}
 
\mypar{What if the number of classes increases?} The gap is even more pronounced on the more difficult Tiny-ImageNet-C testbed, where our $34.9\%$ mean accuracy is $5.6\%$ higher than that of the strongest baseline, BATCLIP. Finally, on the highly challenging 1000-class ImageNet-C benchmark, where many methods show limited gains, \Ours{} again proves its robustness. Specifically, achieves the highest mean accuracy, $26.0\%$, a promising $+5.5\%$ gain over the CLIP baseline, outperforming all other TTA baselines. 

\begin{table}[t]
\centering
\setlength{\tabcolsep}{4.0pt} % Adjust spacing as needed
\scriptsize
\caption{\textbf{Performance comparison under domain shifts}. Results using CLIP ViT-B/32. 
Best method in \textbf{bold}, second best \underline{underlined}.}
\label{tab:alldomains_transposed}
\vspace{-1em}
\begin{tabular}{l ccccc|c}
\toprule
\multirow{2}{*}{\textbf{Method}}& \textbf{PACS} & \textbf{O-Home} & \textbf{VLCS} & \textbf{VisDA-3D} & \textbf{VisDA-YT} & \textbf{Mean} \\
 & \multicolumn{3}{c}{\textit{(texture / style)}} & \multicolumn{2}{c|}{\textit{(simulated / video)}} & \\
\midrule
CLIP & 93.65 & 77.53 & 80.16 & 84.43 & 84.45 & 84.04 \\
TENT & 93.81 & 77.68 & 80.27 & 84.86 & 84.68 & 84.26 \\
TPT & 93.23 & 77.20 & 74.57 & 79.35 & 83.57 & 81.58 \\
CLIPArTT & 93.95 & 77.56 & 80.06 & 85.09 & 84.40 & 84.21 \\
WATT & \underline{94.80} & 78.83 & \textbf{81.14} & \underline{85.36} & \underline{84.69} & \underline{84.96} \\
BATCLIP & 94.52 & \underline{78.90} & \underline{80.78} & 81.97 & 83.60 & 83.95 \\
\cellcolor{gray!20} \textbf{\Ours{} (\textit{Ours})} & \cellcolor{gray!20}\textbf{95.94} & \cellcolor{gray!20}\textbf{80.15} & \cellcolor{gray!20}78.33 & \cellcolor{gray!20}\textbf{90.73} & \cellcolor{gray!20}\textbf{85.44} & \cellcolor{gray!20}\textbf{86.12} \\
\bottomrule
\end{tabular}
\end{table}

\mypar{Performance under additional shifts:} 
~Table~\ref{tab:alldomains_transposed} (VisDA-C) reports the results of \textit{simulated} (3D) and \textit{video} (YT) shifts, which showcase the superiority of \Ours{}. 
More concretely, the differences compared to the second-competitor SoTA method (WATT) in these two datasets are $5.4\%$ and $0.8\%$. Regarding other popular baselines, per-dataset differences can go up to $11\%$, e.g., $11.4\%$ compared to TPT in the 3D domain. It is worth noting also the significant performance drop of the very recent BATCLIP on this benchmark, which underscores the instability of methods that rely too heavily on their own initial predictions in the face of large domain gaps. Furthermore, Table~\ref{tab:alldomains_transposed} results on \textit{texture and style shifts} show that the trend of superior performance is observed in other datasets (i.e., PACS and OfficeHome), whereas performance is degraded in only one out of the five datasets, i.e., VLCS. All considered, \Ours{} emerges as a powerful and reliable TTA approach, where average improvement gains compared to recent methods are $2.2\%$ (BATCLIP), $1.2\%$ (WATT), and $1.9\%$ (CLIPArTT), showcasing its versatility. 

\subsection{In-depth studies}
\label{ablation}

\begin{figure}[t!]
    \centering
    \includegraphics[width=\linewidth]{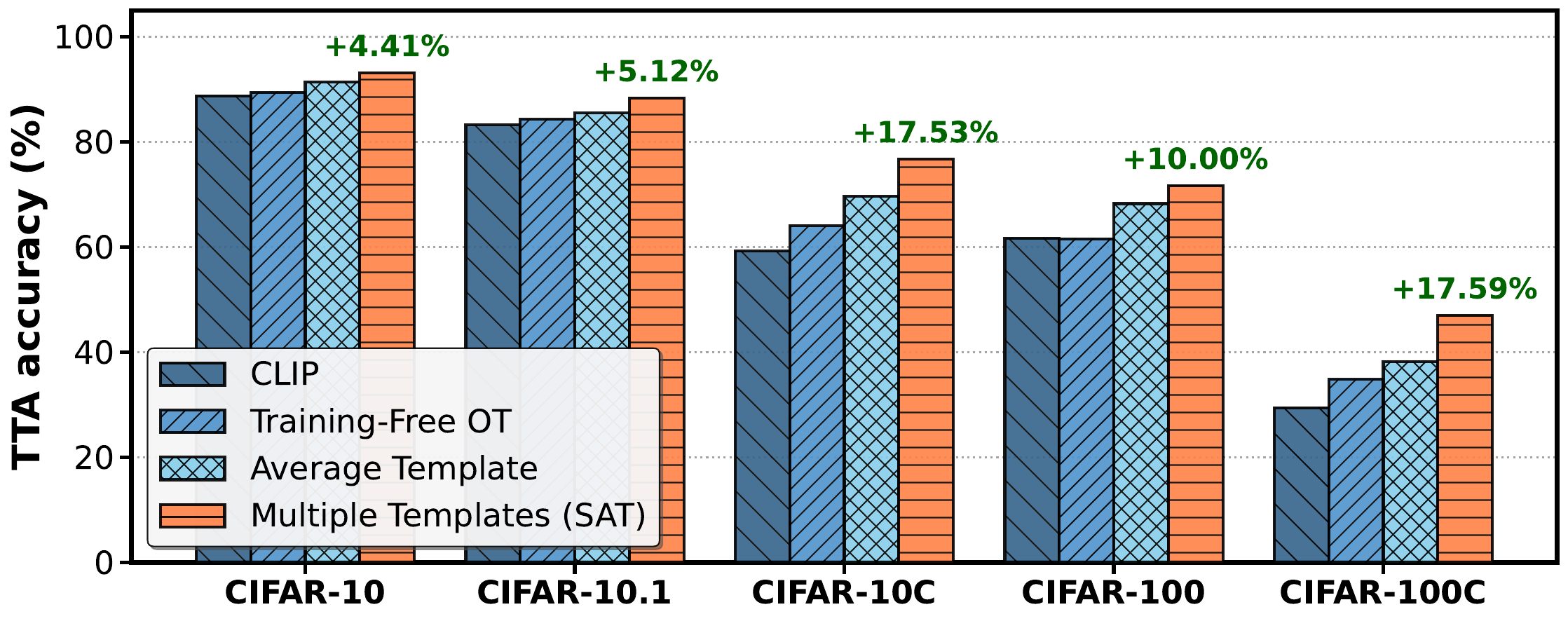}
    \caption{\textbf{Ablation on each component}. Results come from adding each element of \Ours{}. In \textbf{green}, performance differences compared to CLIP (numerical values in \appensecref{ssec:num-values}).
    }
    \label{fig:ablation_components}
    % \vspace{-1.0em}
\end{figure}

\mypar{{Impact of each component.}} In Fig.~\ref{fig:ablation_components}, we empirically validate the benefits of each component in \Ours{} (each configuration is detailed in \appensecref{sec:app_configurations}). 
The ‘Training-Free OT’ (using OT for assignments but not updating the model) already improves over zero-shot CLIP, demonstrating the power of our batch-aware assignment. ‘Average Template’ (using an averaged text prototype) provides a further boost. Finally, our full method, which utilizes ‘Multi-Templates’ knowledge distillation, yields the largest gains, demonstrating that all components are necessary. The benefit of leveraging multiple textual semantic views is also supported in \appensecref{ssec:ablation_template} experiments, where we showcase that performance consistently scales with the number of templates.

\mypar{Do observations hold across backbones?} Tab.~\ref{tab:ViT-B16} reports the performance across different datasets when larger CLIP pre-trained models are employed. These results demonstrate the model-agnostic nature of \Ours{}, as its superiority remains consistent across different backbones.

\begin{table}[h!]
\setlength{\tabcolsep}{2pt}
\centering
\scriptsize % Use a consistent font size
\caption{\textbf{TTA results using larger CLIP ViT backbones.}}
% \textcolor{red}{ADD BATCLIP and ImageNet-C results. ALSO, ADD IMPROVEMENTS IN GREEN.}
\label{tab:ViT-B16}
\vspace{-1em}
\begin{tabular}{c}
\begin{tabular}{lcccc}
\toprule
\textbf{a) \underline{ViT-B/16}} & \textbf{CIFAR-10C} & \textbf{CIFAR-100C} & \textbf{Tiny-IN-C} & \textbf{ImageNet-C} \\
\midrule
CLIP            & 60.15 & 32.01 & 20.92 & 20.89\\
TENT            & 68.00 & 37.90 & 29.78 & 22.79\\
TPT             & 59.75 & 33.73 & 26.96 & 22.35\\
CLIPArTT        & 73.22 & 40.08 & 32.90 & 23.47\\
WATT            & 76.22 & 48.95 & 31.66 & 24.38\\
BATCLIP         & 73.52 & 38.85 & 29.30 & 27.16\\
\cellcolor{gray!20} \textbf{\Ours{}} & \cellcolor{gray!20} \textbf{80.11} (\gaintxt{19.96}) & \cellcolor{gray!20} \textbf{51.24} (\gaintxt{19.23}) & \cellcolor{gray!20} \textbf{37.69} (\gaintxt{16.77}) & \textbf{28.35}\cellcolor{gray!20} (\gaintxt{7.46}) \\
\bottomrule
\end{tabular}

\vspace{2mm} \\

\begin{tabular}{lcccc}
\toprule
\textbf{\textbf{\textbf{b) \underline{ViT-L/14}}}} & \textbf{CIFAR-10C} & \textbf{CIFAR-100C} & \textbf{Tiny-IN-C} & \textbf{ImageNet-C} \\
\midrule
CLIP            & 76.04 & 44.59 & 34.98 & 32.05\\
TENT            & 79.18 & 50.14 & 40.28 & 33.09\\
TPT             & 75.01 & 47.58 & 41.07 & 30.67\\
CLIPArTT        & 78.06 & 52.52 & 42.98 & 34.13\\
WATT            & 80.06 & 54.34 & 43.28 & 36.30\\
BATCLIP         & 83.79 & 48.84 & 35.72 & 37.07\\
\cellcolor{gray!20} \textbf{\Ours{} (\textit{Ours})} & \cellcolor{gray!20} \textbf{86.35} (\gaintxt{10.31}) & \cellcolor{gray!20} \textbf{62.21} (\gaintxt{17.62}) & \cellcolor{gray!20} \textbf{50.36} (\gaintxt{15.38}) & \cellcolor{gray!20} \textbf{38.82} (\gaintxt{6.77}) \\
\bottomrule
\end{tabular}

\\

\end{tabular}
\end{table}

\mypar{Generalization across VLMs.}
\Ours{}'s principles are also model-agnostic beyond the CLIP pre-training framework. As shown in Table~\ref{tab:siglip}, \Ours{} consistently delivers state-of-the-art performance when applied to a modern SigLIP backbone. The gains are even more pronounced on this architecture, with a $15.0\%$ improvement on CIFAR-100C and $5.4\%$ on ImageNet-C over zero-shot SigLIP, proving the general applicability of our cross-modal alignment paradigm.

\begin{table}[t]
\centering
\scriptsize
\setlength{\tabcolsep}{1.7pt}
\caption{\textbf{Performance using SigLIP \cite{zhai2023sigmoidlosslanguageimage}}.}
\label{tab:siglip}
\vspace{-1em}
\begin{tabular}{lcccc}
\toprule
\textbf{Method} & \textbf{CIFAR-10C} & \textbf{CIFAR-100C} & \textbf{Tiny-IN-C} & \textbf{ImageNet-C} \\
\midrule
SigLIP  
   & 59.04
   & 34.76 
   & 22.00
   & 26.49 \\
BATCLIP 
   & 67.45 %(\gaintxt{+8.41})
   & 39.26 %(\gaintxt{+4.50})
   & 24.72 %(\gaintxt{+2.72})
   & 30.40 %(\gaintxt{+3.91}) 
   \\
\cellcolor{gray!20} \textbf{\Ours{} (\textit{Ours})} 
   & \cellcolor{gray!20} \textbf{76.20} (\gaintxt{17.16})
   & \cellcolor{gray!20} \textbf{49.72} (\gaintxt{14.96})
   & \cellcolor{gray!20} \textbf{26.81} (\gaintxt{4.81})
   & \cellcolor{gray!20} \textbf{31.92} (\gaintxt{5.43}) \\
\bottomrule
\end{tabular}
%\vspace{-6pt}
\end{table}

\begin{figure}[h!]
    \centering
    \includegraphics[width=\linewidth]{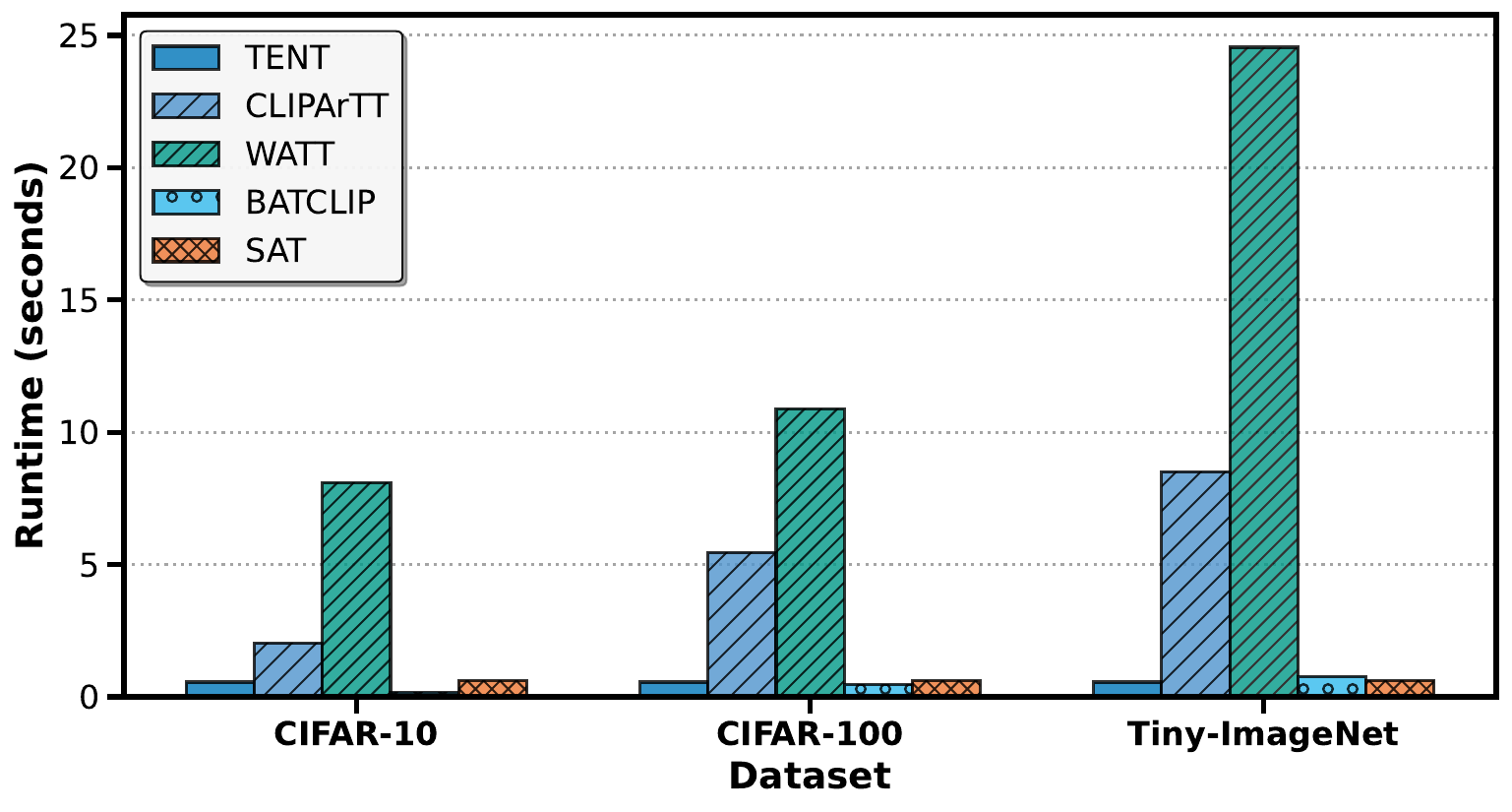}
    \caption{\textbf{Inference Runtime Comparison}. Runtime in seconds for ViT-B/32 TTA methods on an NVIDIA RTX A6000 using a test-time batch of size=128 images.}
    \label{fig:runtimes}
    % \vspace{-1.0em}
\end{figure}

\mypar{Computational analysis.} Fig. \ref{fig:runtimes} depicts the running time required for relevant baselines and the proposed \Ours{} across several datasets, ordered in terms of increasing complexity, i.e., number of categories. These results expose that different TTA methods, particularly SoTA, substantially differ in total runtimes. In particular, recent CLIPArTT and WATT constantly increase the required runtime with the number of classes, driven by their iterative nature, which involves multiple template evaluations and weight averaging during inference. BATCLIP is also highly efficient, yet its runtime still shows a slight increase as the number of classes grows. In contrast, \Ours{} avoids this overhead by distilling this information during the adaptation stage and computing multiple text embeddings \textit{off-line} only once. It is worth noting that the Sinkhorn algorithm used for generating pseudo-codes is highly efficient, as its iterative updates converge faster on the low-dimensional similarity matrices and benefit from parallelizable computations, accounting for only nearly $1\%$ of the total runtime. This combination of speed and consistency establishes \Ours{} as a scalable and effective method, providing runtimes from one to two orders of magnitude lower than CLIPArTT and WATT for the most complex dataset (i.e., Tiny-ImageNet), while significantly outperforming them from a discriminative standpoint.

%% file: 5_Conclusions.tex
\section{Conclusions}
\label{sec:conc}
We have presented Semantic Anchor Transport (\Ours{}), a novel approach for the test-time adaptation of vision-language models. \Ours{} reformulates TTA as a principled cross-modal alignment problem. It generates robust, batch-aware pseudo-labels by aligning visual embeddings to fixed text-based semantic anchors using Optimal Transport. This global assignment strategy fundamentally mitigates the error accumulation demonstrated by other TTA methods. Furthermore, SAT employs a sophisticated multi-template distillation strategy to harness diverse textual clues, enhancing robustness without incurring significant computational overhead. Extensive experiments demonstrate that SAT achieves state-of-the-art performance on TTA across multiple visual domain shift benchmarks and multi-modal backbones.

%and is a computationally efficient solution.

% \mypar{Limitations.} In the context of TTA, the long-tailed (LT) issue is well-known \cite{zhang2022self,zhao2023delta},
% manifesting as an obvious bias in test-time optimization towards dominant classes \cite{zhao2023delta}. The proposed solution integrates an entropic constraint to efficiently solve the optimal transport problem, implicitly enforcing class-balanced solutions. This %behavior 
% enhances the performance on minority classes, which is particularly desirable in LT. Nevertheless, this may come at the cost of degrading the performance on the majority classes, which have a larger weight on the standard figures of merit employed in TTA (e.g., performance on VLCS). Thus, even though we consider that not having evaluated the proposed approach in specific LT scenarios (which is a research topic by itself) is a limitation of this study, \cite{zhang2022self} actually integrates a mechanism to encourage class-balanced predictions to address LT in TTA. We believe that studying the properties of \Ours{} in long-tailed TTA, and exploring alternative solutions to encourage balanced distributions, remains a promising future research direction.

%% file: 6_Appendix.tex
\clearpage
\setcounter{page}{1}
\setcounter{section}{0}
\renewcommand{\thesection}{\Alph{section}}
\maketitlesupplementary

\section{Additional datasets details.}
\label{appendix_dataset}
\mypar{Datasets}.
CIFAR-10.1 introduces a natural shift from CIFAR-10, whereas CIFAR-10-C and CIFAR-100-C are augmented with 15 different corruptions across 5 severity levels (each containing 10,000 images), leading to 75 corruption scenarios commonly employed in domain shift problems. These datasets are critical for evaluating how well models can generalize to real-world variations and noise that are not present in the training set. Similarly, ImageNet-C provides these corruptions for the 1000-class ImageNet validation set. These datasets are critical for evaluating how well models can generalize to real-world variations and noise. Tiny-ImageNet is a downsized version of the original ImageNet dataset, providing a more accessible challenge with 200 classes. Tiny-ImageNet-C further extends Tiny-ImageNet by incorporating various common corruptions. This dataset is instrumental for testing robustness against a variety of distortions. To assess the performance on class imbalanced datasets, we employ several datasets from Domainbed (PACS, OfficeHome, VisDA-C) that are often utilized for benchmarking domain adaptation algorithms, and have been recently adopted in TTA scenarios \cite{hakim2024clipartt}. Each dataset presents unique domain shift challenges and diverse visual categories, allowing for a comprehensive evaluation of a model’s ability to generalize across different environments.
 
\mypar{CLIP's text templates}.
In the experimental setup, several predefined text templates from CLIP \cite{clipad} were used to evaluate the proposed model's adaptability and performance. These are designed to generalize across different contexts and image types, enabling a robust assessment of the model's capabilities in handling diverse visual and textual representations. In Table \ref{tab:templates}, we provide the templates used in our experiments. Each template includes a placeholder ${}$, which is dynamically replaced with the class name during the generation of text prompts. Note that, for the sake of fairness with prior literature, the templates are the same as the ones employed in WATT \cite{osowiechi2024watt}.

\begin{table}[!h]
    \caption{The different templates used during the experiments.}
    \label{tab:templates}
    \scriptsize
    \centering    
    %\dorowcolors   
    \begin{tabular}{ll}
    \toprule
        ~ & Template\\ \midrule
      1: & ``\texttt{a photo of a \{class $k$\}}''\\
      2: & ``\texttt{itap of a \{class $k$\}}'' \\
      3: & ``\texttt{a bad photo of the \{class $k$\}}'' \\
      4: & ``\texttt{a origami \{class $k$\}}'' \\
      5: & ``\texttt{a photo of the large \{class $k$\}}'' \\
      6: & ``\texttt{a \{class $k$\} in a video game}'' \\
      7: & ``\texttt{art of the \{class $k$\}}''\\
      8: & ``\texttt{a photo of the small \{class $k$\}}''\\ \bottomrule
    \end{tabular}
\end{table}

\section{Additional experimental details.}
\label{appendix_experiments}

% This subsection is cited in Sec. 5.3
\subsection{Configuration: Baselines}
\label{sec:app_configurations}

In this section, we detail the configurations used to assess the impact of each main component of our approach, which we refer to \textit{Training-free} OT and \Ours{}  with Average Template. Note that, in short, \textit{Training-free} OT motivates the use of the soft assignments $\mathbf{Q}^*$ over the CLIP baseline, whereas \Ours{} with Average Template includes these assignments to fine-tune the model. Furthermore, we want to highlight that the latter does not fully leverage multiple individual templates, which is introduced in our proposed method. A more detailed description follows. 
Given a batch of test images, \textit{Training-free} OT (Algorithm \ref{algo:ours_trainFree}) computes soft assignments for each text template, which are later averaged to produce a final prediction. 
On the other hand, SAT with Average Template (Algorithm \ref{algo:ours_avg}) utilizes the averaged assignments (obtained by \textit{Training-free} OT) to refine the visual encoder. More concretely, at each batch, the average assignments $\tilde{\mathbf{Q}}$ supervise the predictions of the test images produced by the model. These predictions are obtained by resorting to Eq. (\ref{eq:clip}), where the average class embedding $\mathbf{T}$ is used, together with the visual embeddings of the test images, $\mathbf{Z}$. Differences between the assignments and the predictions are minimized via a cross-entropy loss, whose gradients are used to update the layer norm parameters of the model. Then, once the model is updated, we can do the final inference for the test images.

\begin{algorithm}[ht]
\caption{\textit{Training-free} OT.}
\label{algo:ours_trainFree}
\begin{algorithmic}[1]
    \State \textbf{input:} test dataset $\mathcal{X} = \{\vx_n\}$ (images), text templates~$\mathcal{T}$ (class descriptions), visual and text encoders ($\boldsymbol{\theta}$, $\phi$).

    \noindent \textcolor{gray}{// Split $\mathcal{X}$ into $B$ batches of size $B_T$.}
    
    \noindent \textcolor{gray}{// Compute $K \times M$ text prototypes ($\boldsymbol{\phi}(\mathcal{T}), \forall m, k)$.}
    \State $\T \in \mathbb{R}^{d \times K \times M} = \left[\vt_{km} \right]_{k=1, \dots, K, m=1, \dots, M}$ 
    
    \For{sampled minibatch $\{\vx_i\}_{i=1}^{B_T}$}
        \For{each template $m$ in a random permutation of $\{1, 2, \dots, M\}$}  
            
            \hspace{2mm} \textcolor{gray}{// Step 1: Align - Compute soft assignments $\mathbf{Q}^*_m$.}
            \State $\mathbf{T}_m \in \mathbb{R}^{d \times K}$ \textcolor{gray}{// class text embeddings for $m$.} 
            \State $\mathbf{Z} = [\vz_1, \dots, \vz_{B_T}]$ \textcolor{gray}{// visual features ($\boldsymbol{\theta}_{\text{LN}}^{(m-1)}$).} 
            \State Compute soft assignments $\mathbf{Q}^*_m$ \textcolor{gray}{// Eq. (\ref{eq:individual-codes}).}
        \EndFor
        
        \hspace{-2mm} \textcolor{gray}{// Inference (for all images in the batch).}
        \State $\mathbf{P} = [\vp_1, \dots, \vp_{B_T}]$ \textcolor{gray}{// predict: $\vp_i = \frac{1}{M}\sum_{m=1}^M \vq_{im}$.}

    \EndFor
\end{algorithmic}
\end{algorithm}

\begin{algorithm}[ht]
\caption{\Ours{} with Average Template.}
\label{algo:ours_avg}
\begin{algorithmic}[1]
    \State \textbf{input:} test dataset $\mathcal{X} = \{\vx_n\}$ (images), text templates~$\mathcal{T}$ (class descriptions), visual and text encoders ($\boldsymbol{\theta}$, $\phi$).

    \noindent \textcolor{gray}{// Split $\mathcal{X}$ into $B$ batches of size $B_T$.}
    \noindent \textcolor{gray}{// Compute $K \times M$ text prototypes ($\boldsymbol{\phi}(\mathcal{T}), \forall m, k)$.}
    \State $\T \in \mathbb{R}^{d \times K \times M} = \left[\vt_{km} \right]_{k=1, \dots, K, m=1, \dots, M}$ 
    
    \For{sampled minibatch $\{\vx_i\}_{i=1}^{B_T}$}
        
        \hspace{-2mm} \textcolor{gray}{// 1 - Adaptation}
        \For{each template $m$ in a random permutation of $\{1, 2, \dots, M\}$}  
            
            \hspace{2mm} \textcolor{gray}{// Step 1: Align - Compute soft assignments $\mathbf{Q}^*_m$.}
            \State $\mathbf{T}_m \in \mathbb{R}^{d \times K}$ \textcolor{gray}{// class text embeddings for $m$.} 
            \State $\mathbf{Z} = [\vz_1, \dots, \vz_{B_T}]$ \textcolor{gray}{// visual features ($\boldsymbol{\theta}_{\text{LN}}^{(m-1)}$).} 
            \State Compute soft assignments $\mathbf{Q}^*_m$ \textcolor{gray}{// Eq. (\ref{eq:individual-codes}).}
        \EndFor

        \hspace{-2mm} \textcolor{gray}{// Step 2: Compute average assignments and refine the encoder.}
        \State $\tilde{\mathbf{Q}} = \frac{1}{M}\sum_{m=1}^M \mathbf{Q}^*_m$ \textcolor{gray}{// average assignments.}
        \State $\mathbf{T} = \frac{1}{M}\sum_{m=1}^M \mathbf{T}_m$ \textcolor{gray}{// average class text embeddings.}
        \State $\mathbf{P}= [\vp_1, \dots, \vp_{B_T}]$ \hfill \textcolor{gray}{// predict with $\mathbf{T}$, Eq. (\ref{eq:clip}).}
        \State Min. cross-entropy loss between $\mathbf{P}$ and $\tilde{\mathbf{Q}}$ \textcolor{gray}{// Eq. (\ref{eq:posterior}).}
        \State $\theta_{\text{LN}}^{(t)} \rightarrow \theta_{\text{LN}}^{(t+1)}$ \textcolor{gray}{// update layer norm.}
        
        \hspace{-2mm} \textcolor{gray}{// 2 - Inference (for all images in the batch).}
        \State $\mathbf{Z} = [\vz_1, \dots, \vz_{B_T}]$ \hfill \textcolor{gray}{// visual features (with $\boldsymbol{\theta}_{\text{LN}}^{(m)}$).}
        \State $\mathbf{P}= [\vp_1, \dots, \vp_{B_T}]$ \hfill \textcolor{gray}{// predict with $\mathbf{T}$, Eq. (\ref{eq:clip}).}

    \EndFor
\end{algorithmic}
\end{algorithm}

\subsection{Extended numerical values}
\label{ssec:num-values}

We further substantiate the findings presented in Figure \ref{fig:ablation_components} by providing detailed results in Tables \ref{tab:CIFAR-10C} and \ref{tab:CIFAR-100C}. The tables showcase a comprehensive performance comparison of the impact of the different components of our approach, which empirically motivate our model. These results are reported for different corruption benchmarks on the CIFAR-10, CIFAR-10.1, CIFAR-10C, CIFAR-100, and CIFAR-100C datasets, respectively, using a ViT-B/32 backbone.

\begin{table}[ht]
\caption{Numerical Analysis for CIFAR-10, CIFAR-10.1 and CIFAR-10-C as shown in Figure \ref{fig:ablation_components}.}
\label{tab:CIFAR-10C}
\setlength{\tabcolsep}{3.25pt}
   \centering
   \scriptsize
   \begin{tabular}{llcccc}
       \toprule
       & \textbf{Dataset} & \textbf{CLIP} & \textbf{Training-Free OT} & \textbf{Avg. Template} & \textbf{\Ours} \\ 
       \midrule
       & CIFAR-10      & 88.74 & 89.44 & 91.40 & \textbf{93.15} \\ 
       & CIFAR-10.1    & 83.25 & 84.30 & 85.50 & \textbf{88.37} \\ 
       \midrule
       & Gaussian Noise      & 35.27 & 46.68 & 54.68 & \textbf{64.85} \\ 
       & Shot Noise          & 39.67 & 49.84 & 57.13 & \textbf{67.34} \\ 
       & Impulse Noise       & 42.61 & 47.18 & 52.72 & \textbf{62.27} \\ 
       & Defocus Blur        & 69.76 & 73.22 & 77.84 & \textbf{82.09} \\ 
       & Glass Blur          & 42.40 & 50.09 & 56.88 & \textbf{68.07} \\ 
       & Motion Blur         & 63.97 & 70.15 & 74.81 & \textbf{81.30} \\ 
       & Zoom Blur           & 69.83 & 74.19 & 77.99 & \textbf{83.13} \\ 
       & Snow                & 71.78 & 73.93 & 78.41 & \textbf{83.71} \\ 
       & Frost               & 72.86 & 75.66 & 79.31 & \textbf{83.40} \\ 
       & Fog                 & 67.04 & 69.28 & 75.54 & \textbf{82.56} \\ 
       & Brightness          & 81.87 & 82.90 & 86.87 & \textbf{89.90} \\ 
       & Contrast            & 64.37 & 67.30 & 75.30 & \textbf{84.86} \\ 
       & Elastic Transform   & 60.83 & 64.06 & 69.54 & \textbf{76.08} \\ 
       & Pixelate            & 50.53 & 56.65 & 62.91 & \textbf{76.25} \\ 
       & JPEG Compression    & 55.48 & 59.75 & 64.31 & \textbf{70.03} \\ 
       \midrule
       & \textbf{Mean}       & 59.22 & 64.05 & 69.62 & \textbf{77.06} \\ 
       \bottomrule
   \end{tabular}
\end{table}

\begin{table}[ht]
\caption{Numerical Analysis for CIFAR-100 and CIFAR-100-C datasets as shown in Figure \ref{fig:ablation_components}.}
\label{tab:CIFAR-100C}
\setlength{\tabcolsep}{3.25pt}
\centering
\scriptsize
\begin{tabular}{lcccc}
    \toprule
    \textbf{Corruption} & \textbf{CLIP} & \textbf{Training-Free OT} & \textbf{Avg. Template} & \textbf{\Ours} \\ 
    \midrule
    CIFAR-100      & 61.68  & 61.55 & 68.26 & \textbf{71.68} \\ 
    \midrule
    Gaussian Noise      & 14.80  & 21.26 & 23.29 & \textbf{33.43} \\ 
    Shot Noise          & 16.03  & 22.98 & 25.15 & \textbf{35.60} \\ 
    Impulse Noise       & 13.85  & 22.27 & 22.14 & \textbf{30.94} \\ 
    Defocus Blur        & 36.74  & 41.40 & 46.84 & \textbf{53.87} \\ 
    Glass Blur          & 14.19  & 22.47 & 23.60 & \textbf{35.26} \\ 
    Motion Blur         & 36.14  & 40.71 & 44.41 & \textbf{52.77} \\ 
    Zoom Blur           & 40.24  & 45.14 & 49.97 & \textbf{56.71} \\ 
    Snow                & 38.95  & 44.14 & 46.73 & \textbf{54.30} \\ 
    Frost               & 40.56  & 45.03 & 47.85 & \textbf{54.92} \\ 
    Fog                 & 38.00  & 40.00 & 45.45 & \textbf{53.57} \\ 
    Brightness          & 48.18  & 51.26 & 57.22 & \textbf{64.43} \\ 
    Contrast            & 29.53  & 33.69 & 41.16 & \textbf{55.01} \\ 
    Elastic Transform   & 26.33  & 33.02 & 34.85 & \textbf{43.79} \\ 
    Pixelate            & 21.98  & 27.72 & 30.13 & \textbf{44.51} \\ 
    JPEG Compression    & 25.91  & 31.69 & 33.80 & \textbf{40.83} \\ 
    \midrule
    \textbf{Mean}       & 29.43  & 34.85 & 38.17 & \textbf{47.33} \\ 
    \bottomrule
\end{tabular}
\end{table}

\subsection{Additional Details on Epsilon (\texorpdfstring{$\epsilon$}{epsilon})}
\label{epsilon_study}
\begin{figure}[t!]
    \centering
    \includegraphics[width=0.8\linewidth]{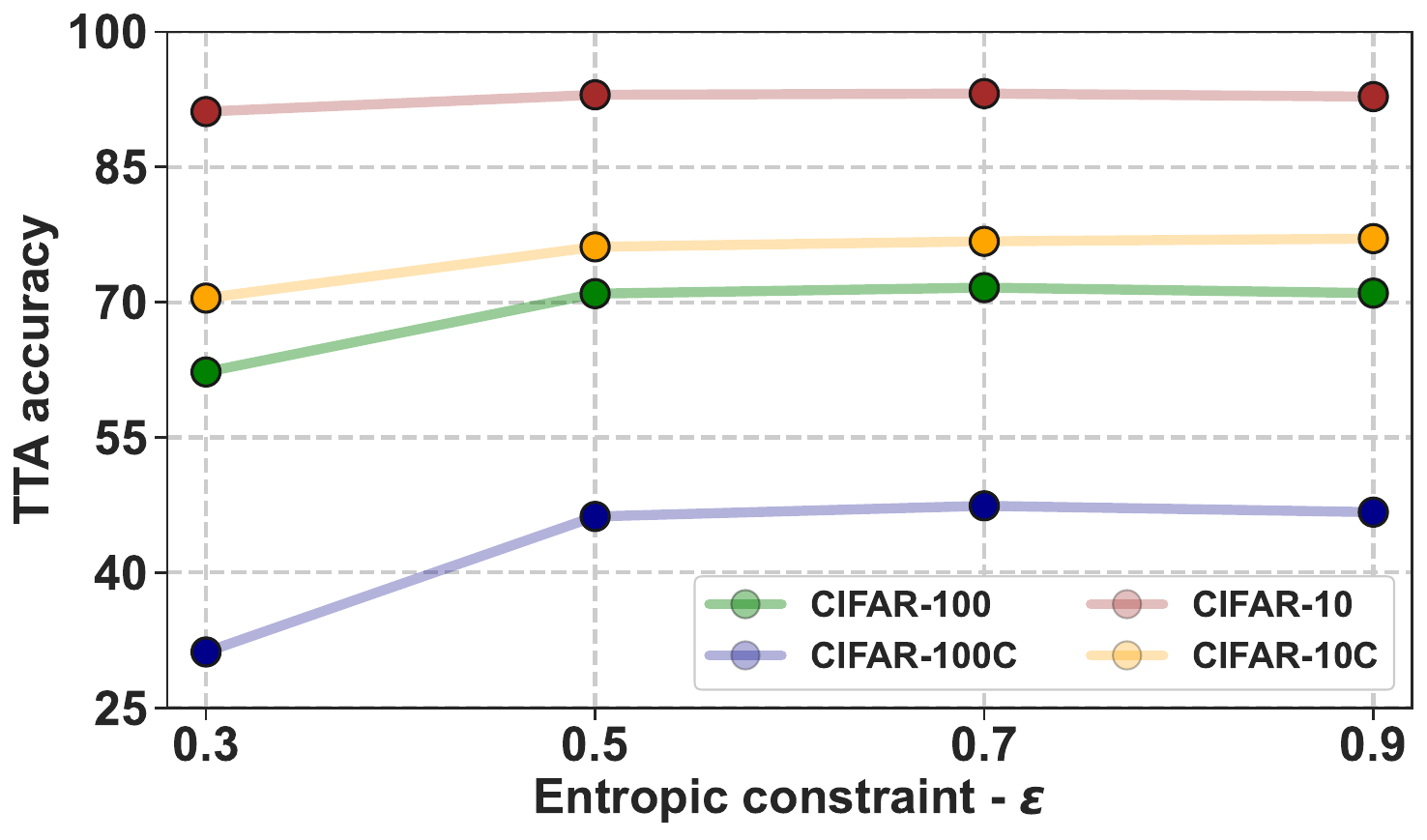}
    \caption{\textbf{Ablation on epsilon values} across multiple datasets.}
    \label{fig:epsilon}
\end{figure}

In the main paper, we stated that $\epsilon$, the entropic constraint weight in Eq. \ref{eq:main} was set to $\epsilon=0.7$ based on preliminary experiments. However, here in the appendix, we explore the sensitivity of the model to different values of $\epsilon$ to assess its impact on accuracy under various corruptions and datasets. As shown in Table \ref{tab:epsilon_comp}, we experimented with $\epsilon$ values of 0.3, 0.5, 0.7 and 0.9 across CIFAR-10C and CIFAR-100C datasets. 
Further, we show the values for CIFAR-10 and CIFAR-100 datasets along with averaged performance over corruptions in the Figure \ref{fig:epsilon}.

\begin{table}[ht]
    \caption{Accuracy on CIFAR-10 and CIFAR-100 under different corruptions and epsilon values.}
    \label{tab:epsilon_comp}
    \centering
    \scriptsize
    \setlength{\tabcolsep}{4pt} % Tighter column spacing (default is 6pt)
        \begin{tabular}{@{}lcccccccc@{}}
        \toprule
        \textbf{Corruption} & \multicolumn{4}{c}{\textbf{CIFAR-10}} & \multicolumn{4}{c}{\textbf{CIFAR-100}} \\ 
        \cmidrule(lr){2-5} \cmidrule(lr){6-9}
                            & \textbf{0.3} & \textbf{0.5} & \textbf{0.7} & \textbf{0.9} & \textbf{0.3} & \textbf{0.5} & \textbf{0.7} & \textbf{0.9} \\ \midrule
        Original            & 91.15        & 93.01        & 93.15        & 92.81        & 62.26        & 70.95        & 71.62        & 71.01        \\
        \midrule
        Gaussian Noise      & 56.16        & 63.65        & 64.43        & 65.26        & 13.74        & 32.03        & 33.2         & 32.44        \\
        Shot Noise          & 57.53        & 66.29        & 66.27        & 67.4         & 15.62        & 34.69        & 35.33        & 35.46        \\
        Impulse Noise       & 54.27        & 61.22        & 61.84        & 62.42        & 19.44        & 29.94        & 30.99        & 30.36        \\
        Defocus Blur        & 76.27        & 81.33        & 82.1         & 82.49        & 39.16        & 52.77        & 54.16        & 53.32        \\
        Glass Blur          & 57.63        & 66.75        & 67.55        & 67.96        & 17.06        & 34.46        & 35.24        & 34.89        \\
        Motion Blur         & 71.99        & 80.11        & 81.31        & 80.9         & 36.59        & 51.33        & 52.67        & 51.62        \\
        Zoom Blur           & 76.24        & 82.54        & 83.11        & 83.46        & 42.58        & 55.53        & 56.93        & 56.13        \\
        Snow                & 78.59        & 83.39        & 82.99        & 83.56        & 40.91        & 53.54        & 54.06        & 54.12        \\
        Frost               & 78.92        & 82.67        & 82.93        & 83.41        & 40.98        & 54.1         & 54.93        & 54.37        \\
        Fog                 & 75.75        & 81.64        & 81.9         & 82.47        & 40.76        & 52.69        & 53.6         & 53.12        \\
        Brightness          & 85.97        & 89.59        & 89.72        & 89.71        & 48.31        & 63.27        & 64.87        & 63.57        \\
        Contrast            & 70.38        & 83.55        & 84.53        & 84.6         & 32.09        & 53.61        & 55.39        & 54.41        \\
        Elastic Transform   & 69.17        & 75.42        & 75.67        & 76.09        & 27.78        & 42.69        & 43.78        & 43.01        \\
        Pixelate            & 61.31        & 74.57        & 76.51        & 76.08        & 22.79        & 42.64        & 44.96        & 43.18        \\
        JPEG Compression    & 66.02        & 69.42        & 70.46        & 70.03        & 30.16        & 40.16        & 41.02        & 40.67        \\ 
        \midrule
        Mean                & 70.46        & 76.14        & 76.75        & 77.06        & 31.20        & 46.23        & 47.41        & 46.71        \\ 
        \bottomrule
        \end{tabular}
\end{table}

We can observe that higher $\epsilon$ values (0.7 and 0.9) improved stability and performance due to stronger regularization introduced by higher entropic constraints, which prevents degenerate solutions during Sinkhorn normalization. While smaller $\epsilon$ values sometimes performed better under mild corruptions, they exhibited significant instability during Sinkhorn normalization. In several cases, the $\Q$ matrix became NaN, especially under high noise corruptions, rendering the adaptation process unusable which can be attributed to insufficient regularization, causing numerical instabilities during iterative updates. 
The chosen value ($\epsilon=0.7$) strikes a balance between stability and performance, yielding robust results across both clean and corrupted datasets.

\subsection{Sensitivity to Batch Size}
\label{sec:app_batch_size}
Our method, SAT, is fundamentally a \textit{batch-aware} approach. We evaluate \Ours's performance on CIFAR-10C while varying the batch size from 16 to 128, comparing it against other methods. As shown in Table \ref{tab:batch_size_ablation}, \Ours's performance is robust and scales well, achieving strong results even with a batch size of 32. Notably, under distribution shift on CIFAR-10C, \Ours exhibits substantially lower performance degradation when reducing the batch size (only –2.5\%, from 77.35 to 74.40) compared to much larger drops observed for WATT (–7.5\%) and BATCLIP (–6\%), highlighting \Ours's superior robustness to batch size constraints in corrupted settings. Performance for all methods generally improves with a larger batch, confirming that a sufficiently large batch is important for the stability of TTA.

\begin{table}[h!]
\centering
\small
\caption{Accuracy (\%) on \textbf{CIFAR-10} and \textbf{CIFAR-10C} vs. \textbf{Batch Size}. Results are on ViT-B/32.}
\label{tab:batch_size_ablation}
\begin{tabular}{llcccc}
\toprule
\textbf{Dataset} & \textbf{Method} & \textbf{16} & \textbf{32} & \textbf{64} & \textbf{128} \\
\midrule
\multirow{3}{*}{CIFAR-10} & WATT & \textbf{89.14} & 89.51 & 90.16 & 91.05 \\
 & BATCLIP & 84.31 & 85.22 & 88.34 & 90.35 \\
 & \textbf{SAT (Ours)} & 87.06 & \textbf{92.34} & \textbf{93.05} & \textbf{93.22} \\
\midrule
\multirow{3}{*}{CIFAR-10C} & WATT & 65.66 & 68.34 & 71.21 & 73.00 \\
 & BATCLIP & 63.08 & 61.89 & 65.51 & 69.61 \\
 & \textbf{SAT (Ours)} & \textbf{74.40} & \textbf{77.01} & \textbf{77.35} & \textbf{77.06} \\
\bottomrule
\end{tabular}
\end{table}

% \subsection{Analysis of Catastrophic Forgetting}
% \label{sec:app_forgetting}

% A critical challenge in adaptation is catastrophic forgetting, where a model, in adapting to a new distribution (e.g., corrupted data), loses its knowledge of the original source domain. We evaluate this by testing for performance on the clean data after adapting to each corruption individually. For each of the 15 corruptions in CIFAR-10C and CIFAR-100C, we first adapt the model parameters to that specific corruption. We then evaluate this adapted model on the original, clean test set (CIFAR-10 or CIFAR-100). The accuracy values reported in Table \ref{tab:forgetting} are the means of those 15 clean-set evaluations. As shown, \Ours{} does not suffer from catastrophic forgetting. Adapting to the corrupted distributions improves performance on the original clean data, with a gain of $+2.08$ on CIFAR-10 and a substantial $+5.58$ on CIFAR-100. This suggests \Ours learns a more robust and generalized representation rather than simply overfitting to the corruptions.

% \begin{table}[h!]
% \centering
% \small
% \caption{Catastrophic forgetting analysis. We report accuracy (\%) on the \textbf{clean} test sets after adapting on the full \textbf{corrupted} datasets.}
% \label{tab:forgetting}
% \begin{tabular}{lccc}
% \toprule
% \textbf{Dataset} & \textbf{Original (Clean)} & \textbf{After Adaptation} & \textbf{Gain} \\
% \midrule
% CIFAR-10 & 88.74 & 90.82 & \gaintxt{+2.08} \\
% CIFAR-100 & 61.68 & 67.26 & \gaintxt{+5.58} \\
% \bottomrule
% \end{tabular}
% \end{table}

% This subsection is cited in Sec. 5.3
\subsection{On the impact of multiple templates.}
\label{ssec:ablation_template}
We now assess how the performance evolves as the number of templates increases (Table \ref{tab:clipot_template_performance}). Concretely, we trained our model multiple times, varying the number of templates $N \in \{1,\dots,8\}$. For each $N$, the average accuracy over all template combinations was computed, leading to 256 experiments per dataset. Results from this experiment show that \Ours{} performance consistently increases with the number of templates. This showcases how leveraging multiple templates individually can extract richer information to adapt CLIP at test time.

\begin{table}[ht!]
\centering
\scriptsize  % Use a consistent font size
\caption{Accuracy (\%) of \Ours{} with varying numbers of templates (see Table \ref{tab:templates}) on natural domain shift benchmarks.}
\label{tab:clipot_template_performance}
\begin{tabular}{>{\raggedright\arraybackslash}p{1.6cm}>{\centering\arraybackslash}p{.35cm}>{\centering\arraybackslash}p{.35cm}>{\centering\arraybackslash}p{.35cm}>{\centering\arraybackslash}p{.35cm}>{\centering\arraybackslash}p{.35cm}>{\centering\arraybackslash}p{.35cm}>{\centering\arraybackslash}p{.35cm}>{\centering\arraybackslash}p{.35cm}}

    \toprule
    \textbf{$\#$ Templates} & \textbf{1} & \textbf{2} & \textbf{3} & \textbf{4} & \textbf{5} & \textbf{6} & \textbf{7} & \textbf{8} \\
    \midrule
    \textbf{CIFAR-10}       & 91.59         & 92.16         & 92.46         & 92.69         & 92.88         & 93.04         & 93.14         & 93.14         \\
    \textbf{CIFAR-100}      & 68.36         & 69.46         & 70.08         & 70.56         & 70.90         & 71.15         & 71.38         & 71.34         \\
    \textbf{Tiny-ImageNet}  & 60.62         & 61.73         & 62.32         & 62.69         & 62.95         & 63.16         & 63.20         & 63.11         \\
    \bottomrule
\end{tabular}
\end{table}

\section{Detailed Experimental Results}
\label{sec:app_detailed_results}
To verify that our adaptation method does not simply overfit to corruptions at the cost of performance on the original data distribution, we evaluate SAT on clean, non-corrupted benchmarks. Table \ref{tab:vitb32_clean_batclip} presents the performance of SAT and competing baselines on the standard CIFAR-10, CIFAR-10.1, CIFAR-100, and Tiny-ImageNet test sets.

\subsection{Performance on Clean Datasets}
\label{sec:app_clean}
Table \ref{tab:vitb32_clean_batclip} details the performance of all ViT-B/32 methods on the original, uncorrupted test sets. These results correspond to the clean dataset performance reported in the ablations (Figure \ref{fig:ablation_components}) and form the basis for the corruption-free evaluation. The results show that \Ours{} does not just preserve performance but consistently improves accuracy over the CLIP baseline and other TTA methods, even without a significant domain shift. On CIFAR-10, SAT achieves $93.15\%$ accuracy, surpassing WATT at $91.05\%$. This trend is even more pronounced in more complex datasets. On CIFAR-100, \Ours{} achieves $71.68\%$, a full 10-point gain over the $61.68\%$ from CLIP and is notably higher than all other baselines. Similarly, on Tiny-ImageNet, \Ours{} $63.69\%$ provides a clear improvement over both the baseline ($58.29\%$) and WATT ($61.35\%$).

\begin{table}[ht]
\centering
\scriptsize % Changed from \small
\setlength{\tabcolsep}{3pt} % Reduced from 5pt
\renewcommand{\arraystretch}{1.1}
\caption{Performance on \textbf{Clean Datasets} (ViT-B/32). Best method in \textbf{bold}.}
\label{tab:vitb32_clean_batclip}
\begin{tabular}{lccccccc}
\toprule
\textbf{Dataset} & \textbf{CLIP} & \textbf{TENT} & \textbf{TPT} & \textbf{CLIPArTT} & \textbf{WATT} & \textbf{BATCLIP} & \textbf{\Ours} \\
\midrule
CIFAR-10 & 88.74 & 91.69 & 88.06 & 90.04 & 91.05 & 88.89 & \textbf{93.15} \\
CIFAR-10.1 & 83.25 & 87.60 & 81.80 & 86.35 & 86.98 & 83.70 & \textbf{88.37} \\
CIFAR-100 & 61.68 & 69.74 & 63.78 & 69.79 & 70.74 & 63.94 & \textbf{71.68} \\
Tiny-IN & 58.29 & 57.72 & 58.90 & 59.85 & 61.35 & 58.93 & \textbf{63.69} \\
\bottomrule
\end{tabular}
\end{table}

\subsection{Per-Corruption Results}
\label{sec:app_corruption_detailed}
To supplement the mean accuracy results presented in the main paper (Table \ref{tab:unified_main}), this section provides a granular, per-corruption performance breakdown. This analysis demonstrates that \Ours{}'s strong average performance stems from a consistent robustness across a wide variety of distribution shifts.

We extend this detailed analysis to the larger-scale backbones, ViT-B/16 (Table \ref{tab:unified_stacked_all_b16}) and ViT-L/14 (Table \ref{tab:unified_stacked_all_l14}).

On ViT-B/16 (Table \ref{tab:unified_stacked_all_b16}): \Ours{} achieves the highest accuracy on all 15 corruption types for both CIFAR-10C and CIFAR-100C. For example, on CIFAR-10C \textit{Pixelate} corruption, \Ours{} ($78.73\%$) significantly outperforms the baseline, WATT ($75.67\%$). The gains are similarly strong on ImageNet-C, where \Ours{} (mean $28.35\%$) consistently outperforms all other methods.

On ViT-L/14 (\ref{tab:unified_stacked_all_l14}): The performance gap widens further, demonstrating \Ours{}'s superior scalability. On CIFAR-100C, \Ours{} (mean $69.48\%$) again dominates every single corruption type. The improvement is validated on challenging corruptions like \textit{Glass Blur}, where \Ours{} ($48.19\%$) more than doubles the performance of the zero-shot baseline ($23.46\%$) and is far ahead of the next-best method, WATT ($33.54\%$). On the difficult Tiny-ImageNet-C, \Ours{}'s mean accuracy of $50.36\%$ is a major leap from the $34.98\%$ baseline and all other TTA methods.

This comprehensive, per-corruption breakdown confirms that SAT provides a fundamentally more robust adaptation mechanism that is effective against diverse types of data shifts, and that these benefits are amplified when applied to larger, more capable vision-language models.

\begin{table*}[h!]
\setlength{\tabcolsep}{3pt} % Adjust column spacing
\centering
\footnotesize % Use font size instead of \resizebox
\caption{\textbf{Unified corruption robustness results} (ViT-B/16, severity = 5). 
Top-1 accuracy (\%) on CIFAR-10C, CIFAR-100C, Tiny-ImageNet-C, and ImageNet-C. 
Best method in \textbf{bold} (with gray cell), second best \underline{underlined}.}
\label{tab:unified_stacked_all_b16}
\begin{tabular}{
lccccccccccccccc|c
}
\toprule
\textbf{Method} &
Gaussian &
Shot &
Impulse &
Defocus &
Glass &
Motion &
Zoom &
Snow &
Frost &
Fog &
Bright &
Contrast &
Elastic &
Pixelate &
JPEG &
\textbf{Mean $\uparrow$} \\
\midrule
\multicolumn{17}{l}{\textbf{CIFAR-10C}} \\
\cmidrule(r){1-17}
CLIP & 37.75 & 41.10 & 51.71 & 70.07 & 42.24 & 65.81 & 72.50 & 73.23 & 76.52 & 68.35 & 83.36 & 61.90 & 53.16 & 48.48 & 56.05 & 60.15 \\
TENT & 31.04 & 40.54 & 58.03 & 77.57 & 47.16 & 76.16 & 79.64 & 81.68 & 83.22 & 80.78 & \underline{89.85} & 79.24 & 62.54 & 67.08 & 65.42 & 68.00 \\
% SAR & 43.60 & 45.15 & 56.73 & 73.17 & 40.44 & 72.89 & 75.86 & 75.04 & 77.26 & 76.67 & 84.57 & 72.99 & 56.73 & 60.35 & 59.82 & 64.75 \\
% VTE & 37.10 & 42.10 & 55.20 & 64.10 & 39.00 & 61.90 & 67.30 & 67.50 & 69.80 & 68.20 & 77.00 & 54.80 & 57.10 & 49.70 & 53.10 & 57.59 \\
TPT & 35.35 & 41.03 & 54.86 & 70.29 & 37.86 & 67.43 & 72.91 & 72.98 & 75.87 & 69.13 & 83.67 & 62.16 & 51.26 & 44.65 & 56.73 & 59.75 \\
WATT & \underline{65.57} & \underline{68.67} & \underline{70.39} & 79.90 & \underline{61.62} & 79.02 & 81.10 & 82.54 & 83.46 & \underline{81.88} & 89.10 & \underline{83.79} & \underline{70.93} & \underline{75.67} & \underline{69.65} & \underline{76.22} \\
CLIPArTT & 60.89 & 65.19 & 67.55 & 78.92 & 57.18 & 76.59 & 79.62 & 81.13 & 81.24 & 78.47 & 88.66 & 75.15 & 69.49 & 71.80 & 66.42 & 73.22 \\
BATCLIP & 61.81 & 65.49 & 64.32 & \underline{80.03} & 53.82 & \underline{79.67} & \underline{81.42} & \underline{82.63} & \underline{83.60} & 80.84 & 87.20 & 81.17 & 69.44 & 63.17 & 68.21 & 73.52 \\
\textbf{Ours} & \cellcolor{gray!20}\textbf{68.50} & \cellcolor{gray!20}\textbf{71.24} & \cellcolor{gray!20}\textbf{76.97} & \cellcolor{gray!20}\textbf{83.33} & \cellcolor{gray!20}\textbf{68.33} & \cellcolor{gray!20}\textbf{83.13} & \cellcolor{gray!20}\textbf{85.35} & \cellcolor{gray!20}\textbf{86.65} & \cellcolor{gray!20}\textbf{86.37} & \cellcolor{gray!20}\textbf{85.60} & \cellcolor{gray!20}\textbf{92.64} & \cellcolor{gray!20}\textbf{86.92} & \cellcolor{gray!20}\textbf{75.18} & \cellcolor{gray!20}\textbf{78.73} & \cellcolor{gray!20}\textbf{72.65} & \cellcolor{gray!20}\textbf{80.11} \\
\midrule
\multicolumn{17}{l}{\textbf{CIFAR-100C}} \\
\cmidrule(r){1-17}
CLIP & 15.88 & 17.49 & 21.43 & 40.10 & 13.48 & 39.82 & 45.45 & 42.77 & 45.39 & 38.98 & 52.55 & 33.32 & 24.39 & 21.89 & 27.21 & 32.01 \\
TENT & 12.28 & 15.07 & 13.13 & 50.35 & 4.84 & 49.85 & 54.76 & 52.38 & 51.66 & 50.74 & 64.26 & 48.69 & 33.56 & 36.20 & 30.80 & 37.90 \\
% SAR & 27.36 & 26.66 & 31.55 & 44.24 & 21.97 & 44.26 & 49.38 & 48.96 & 50.24 & 47.57 & 57.75 & 43.72 & 31.22 & 33.07 & 34.97 & 39.53 \\
% VTE & 17.40 & 20.00 & 24.00 & 34.90 & 19.70 & 34.80 & 39.80 & 46.50 & 47.00 & 40.20 & 48.80 & 26.40 & 28.80 & 25.10 & 25.90 & 31.95 \\
TPT & 15.43 & 16.88 & 22.12 & 41.08 & 18.43 & 40.85 & 46.77 & 47.24 & 48.61 & 39.92 & 55.83 & 33.13 & 27.36 & 21.26 & 30.97 & 33.73 \\
WATT & \underline{35.95} & \underline{37.96} & \underline{44.62} & \underline{53.80} & \underline{33.39} & \underline{52.72} & \underline{57.51} & \underline{56.73} & \underline{56.48} & \underline{53.83} & \underline{66.67} & \underline{55.06} & \underline{40.37} & \underline{47.02} & \underline{42.13} & \underline{48.95} \\
CLIPArTT & 19.01 & 20.27 & 17.66 & 49.86 & 18.34 & 50.00 & 54.13 & 52.80 & 49.56 & 49.92 & 63.76 & 47.86 & 32.93 & 39.49 & 35.56 & 40.08 \\
BATCLIP & 21.79 & 28.07 & 30.81 & 45.46 & 21.16 & 45.04 & 49.82 & 49.52 & 47.81 & 47.32 & 59.12 & 43.72 & 30.89 & 30.17 & 32.06 & 38.85 \\
\textbf{Ours} & \cellcolor{gray!20}\textbf{38.80} & \cellcolor{gray!20}\textbf{40.89} & \cellcolor{gray!20}\textbf{46.39} & \cellcolor{gray!20}\textbf{55.50} & \cellcolor{gray!20}\textbf{37.89} & \cellcolor{gray!20}\textbf{55.10} & \cellcolor{gray!20}\textbf{58.77} & \cellcolor{gray!20}\textbf{57.84} & \cellcolor{gray!20}\textbf{58.09} & \cellcolor{gray!20}\textbf{55.68} & \cellcolor{gray!20}\textbf{67.76} & \cellcolor{gray!20}\textbf{58.29} & \cellcolor{gray!20}\textbf{44.32} & \cellcolor{gray!20}\textbf{49.46} & \cellcolor{gray!20}\textbf{43.87} & \cellcolor{gray!20}\textbf{51.24} \\
\midrule
\multicolumn{17}{l}{\textbf{Tiny-ImageNet-C}} \\
\cmidrule(r){1-17}
CLIP & 4.80 & 6.44 & 4.36 & 22.23 & 7.50 & 33.46 & 29.86 & 31.64 & 33.26 & 24.26 & 40.48 & 1.72 & 23.85 & 22.35 & 27.65 & 20.92 \\
TENT & 13.44 & 17.46 & 9.09 & 32.33 & 12.47 & 44.17 & 39.24 & 38.91 & 42.04 & 31.06 & 49.26 & 3.06 & 34.57 & 37.73 & 41.92 & 29.78 \\
% SAR & 9.22 & 11.78 & 5.37 & 7.10 & 2.62 & 16.02 & 14.14 & 22.88 & 26.47 & 8.23 & 36.64 & 0.60 & 13.24 & 15.98 & 27.95 & 14.55 \\
% VTE & 19.50 & 23.50 & 9.50 & 11.00 & 2.50 & 31.50 & 28.49 & 40.00 & 43.50 & 18.00 & 46.00 & 0.50 & 30.00 & 32.00 & 41.50 & 22.71 \\
TPT & 10.53 & 12.90 & 6.57 & 29.16 & 10.88 & 39.91 & 36.62 & 38.66 & 40.95 & 29.22 & 47.78 & 3.58 & 32.11 & 29.02 & 36.51 & 26.96 \\
WATT & 16.78 & 19.27 & 10.42 & 33.33 & 16.26 & 44.21 & 40.26 & 42.31 & \underline{44.83} & 34.08 & \underline{52.16} & 5.43 & 35.11 & 37.47 & 42.93 & 31.66 \\
CLIPArTT & \underline{19.25} & \underline{21.80} & \underline{13.46} & \underline{33.49} & \underline{18.33} & \underline{45.17} & \underline{40.53} & \underline{42.71} & 44.28 & \underline{34.78} & 51.61 & \underline{7.24} & \underline{35.16} & \underline{41.11} & \underline{44.58} & \underline{32.90} \\
BATCLIP & 11.92 & 15.36 & 10.14 & 31.77 & 14.80 & 43.38 & 39.13 & 38.92 & 38.98 & 31.88 & 49.11 & 5.54 & 32.91 & 36.72 & 38.98 & 29.30 \\
\textbf{Ours} & \cellcolor{gray!20}\textbf{24.88} & \cellcolor{gray!20}\textbf{27.99} & \cellcolor{gray!20}\textbf{20.69} & \cellcolor{gray!20}\textbf{37.27} & \cellcolor{gray!20}\textbf{22.72} & \cellcolor{gray!20}\textbf{48.75} & \cellcolor{gray!20}\textbf{45.16} & \cellcolor{gray!20}\textbf{45.74} & \cellcolor{gray!20}\textbf{48.09} & \cellcolor{gray!20}\textbf{41.21} & \cellcolor{gray!20}\textbf{55.75} & \cellcolor{gray!20}\textbf{13.06} & \cellcolor{gray!20}\textbf{41.69} & \cellcolor{gray!20}\textbf{45.83} & \cellcolor{gray!20}\textbf{46.55} & \cellcolor{gray!20}\textbf{37.69} \\
\midrule
\multicolumn{17}{l}{\textbf{ImageNet-C}} \\
\cmidrule(r){1-17}
CLIP & 10.70 & 11.28 & 10.66 & 19.60 & 13.40 & 18.06 & 17.54 & 26.78 & 27.08 & 32.86 & 49.02 & 14.98 & 11.94 & 30.00 & 29.38 & 20.89 \\
TENT & 7.98 & 8.68 & 8.82 & 21.64 & 17.08 & 21.38 & 19.78 & 28.26 & 28.40 & 33.86 & 49.54 & 18.10 & 13.18 & 32.56 & 32.54 & 22.79 \\
% SAR & 15.96 & 17.46 & 16.60 & 21.24 & \underline{20.42} & 25.14 & 22.20 & 30.44 & 29.46 & 35.84 & 50.32 & 20.94 & 14.66 & 32.82 & 34.60 & 26.26 \\
% VTE & 8.00 & 11.00 & 8.00 & 23.00 & 13.00 & 22.00 & 21.00 & 23.00 & 26.00 & 33.00 & 50.00 & 14.00 & 13.00 & 38.00 & 28.00 & 22.07 \\
TPT & 7.15 & 6.23 & 7.03 & 20.11 & 11.24 & 21.36 & 23.19 & 26.04 & \underline{31.17} & 32.01 & 50.12 & 21.05 & 8.13 & 39.22 & 33.15 & 22.35 \\
WATT & 7.68 & 8.80 & 9.14 & 23.70 & 18.36 & 24.88 & 22.54 & 29.56 & 29.36 & 35.52 & \underline{50.82} & 21.44 & 14.60 & \underline{34.90} & 34.40 & 24.38 \\
CLIPArTT & \underline{19.10} & \underline{19.80} & \underline{19.10} & 14.10 & 18.50 & 19.40 & 17.00 & 26.40 & 29.00 & 35.20 & 47.40 & 2.20 & \textbf{22.20} & 28.10 & 34.65 & 23.47 \\
BATCLIP & 17.64 & 19.00 & 16.50 & \underline{24.46} & \underline{19.04} & \textbf{27.20} & \underline{24.20} & \underline{31.90} & 29.74 & \underline{37.46} & \textbf{52.20} & \underline{21.56} & 18.82 & 32.12 & \textbf{35.56} & \underline{27.16} \\
\textbf{Ours} & \cellcolor{gray!20}\textbf{19.80} & \cellcolor{gray!20}\textbf{21.14} & \cellcolor{gray!20}\textbf{19.20} & \cellcolor{gray!20}\textbf{24.78} & \cellcolor{gray!20}\textbf{21.38} & \cellcolor{gray!20}\underline{26.82} & \cellcolor{gray!20}\textbf{24.40} & \cellcolor{gray!20}\textbf{32.04} & \cellcolor{gray!20}\textbf{32.36} & \cellcolor{gray!20}\textbf{38.54} & \cellcolor{gray!20}49.44 & \cellcolor{gray!20}\textbf{24.42} & \cellcolor{gray!20}\underline{20.16} & \cellcolor{gray!20}\textbf{35.92} & \cellcolor{gray!20}\underline{34.84} & \cellcolor{gray!20}\textbf{28.35} \\
\bottomrule
\end{tabular}
\end{table*}

\begin{table*}[h!]
\setlength{\tabcolsep}{3pt} % Adjust column spacing
\centering
\footnotesize % Use font size instead of \resizebox
\caption{\textbf{Unified corruption robustness results} (ViT-L/14, severity = 5). 
Top-1 accuracy (\%) on CIFAR-10C, CIFAR-100C, Tiny-ImageNet-C, and ImageNet-C. 
Best method in \textbf{bold} (with gray cell), second best \underline{underlined}.}
\label{tab:unified_stacked_all_l14}
\begin{tabular}{
lccccccccccccccc|c
}
\toprule
\textbf{Method} &
Gaussian &
Shot &
Impulse &
Defocus &
Glass &
Motion &
Zoom &
Snow &
Frost &
Fog &
Bright &
Contrast &
Elastic &
Pixelate &
JPEG &
\textbf{Mean $\uparrow$} \\
\midrule
\multicolumn{17}{l}{\textbf{CIFAR-10C}} \\
\cmidrule(r){1-17}
CLIP & 64.64 & 67.82 & 78.21 & 80.73 & 50.29 & 80.75 & 82.75 & 83.01 & 84.90 & 78.44 & 91.67 & 84.20 & 65.45 & 75.10 & 72.58 & 76.04 \\
TENT & 68.87 & 71.95 & 80.22 & 83.10 & 57.12 & 82.69 & 84.91 & 85.99 & 87.15 & 81.30 & 93.07 & 87.93 & 69.96 & 78.88 & 75.49 & 79.18 \\
% SAR & & & & & & & & & & & & & & & & \\
% VTE & & & & & & & & & & & & & & & & \\
TPT & 64.44 & 66.81 & 76.46 & 79.01 & 49.64 & 78.85 & 82.32 & 82.69 & 84.63 & 77.56 & 90.94 & 82.88 & 64.81 & 77.89 & 71.18 & 75.01 \\
WATT & 72.73 & 74.60 & 80.95 & 83.15 & 62.35 & 82.61 & 85.44 & 85.61 & 86.88 & 81.79 & 92.59 & 87.38 & 71.25 & 77.67 & 75.84 & 80.06 \\
CLIPArTT & 70.04 & 71.44 & 79.42 & 81.75 & 58.13 & 80.76 & 83.39 & 84.48 & 85.21 & 79.27 & 91.87 & 86.19 & 67.43 & 77.88 & 74.46 & 78.06 \\
BATCLIP & \underline{75.04} & \underline{79.39} & \underline{84.19} & \underline{86.93} & \underline{71.01} & \underline{86.80} & \underline{87.56} & \underline{89.49} & \underline{89.44} & \underline{86.88} & \underline{94.04} & \underline{92.00} & \underline{79.42} & \underline{84.38} & \underline{80.32} & \underline{83.79} \\
\textbf{Ours} & \cellcolor{gray!20}\textbf{80.47} & \cellcolor{gray!20}\textbf{82.64} & \cellcolor{gray!20}\textbf{87.80} & \cellcolor{gray!20}\textbf{90.33} & \cellcolor{gray!20}\textbf{77.72} & \cellcolor{gray!20}\textbf{88.93} & \cellcolor{gray!20}\textbf{91.69} & \cellcolor{gray!20}\textbf{91.56} & \cellcolor{gray!20}\textbf{91.43} & \cellcolor{gray!20}\textbf{90.82} & \cellcolor{gray!20}\textbf{95.86} & \cellcolor{gray!20}\textbf{94.53} & \cellcolor{gray!20}\textbf{82.16} & \cellcolor{gray!20}\textbf{87.13} & \cellcolor{gray!20}\textbf{82.60} & \cellcolor{gray!20}\textbf{86.35} \\
\midrule
\multicolumn{17}{l}{\textbf{CIFAR-100C}} \\
\cmidrule(r){1-17}
CLIP & 30.55 & 34.58 & 44.89 & 48.88 & 23.46 & 50.83 & 56.02 & 49.03 & 53.27 & 48.51 & 60.53 & 50.24 & 35.07 & 43.86 & 39.11 & 44.59 \\
TENT & 36.93 & 38.23 & 49.09 & 55.23 & 27.02 & 56.03 & 61.19 & 55.60 & 58.21 & 49.26 & 67.34 & 59.91 & 38.49 & 48.37 & 43.44 & 50.14 \\
TPT & 36.10 & 40.96 & 49.69 & 50.43 & 24.35 & 51.94 & 56.96 & 54.89 & 58.15 & 53.37 & 66.60 & 53.64 & 35.72 & 44.32 & 44.42 & 47.58 \\
WATT & \underline{44.13} & \underline{46.63} & \underline{56.26} & \underline{57.66} & \underline{33.54} & \underline{57.81} & \underline{62.74} & \underline{61.04} & \underline{62.76} & \underline{54.70} & \underline{71.60} & \underline{63.95} & \underline{41.27} & \underline{51.22} & \underline{49.78} & \underline{54.34} \\
CLIPArTT & 41.46 & 44.27 & 51.44 & 56.55 & 30.47 & 56.98 & 62.56 & 58.81 & 60.38 & 54.38 & 69.63 & 63.39 & 39.57 & 50.45 & 47.45 & 52.52 \\
BATCLIP & 39.87 & 43.50 & 50.11 & 51.55 & 30.95 & 51.03 & 55.98 & 53.77 & 54.53 & 49.08 & 65.52 & 59.07 & 34.07 & 47.96 & 45.61 & 48.84 \\
\textbf{Ours} & \cellcolor{gray!20}\textbf{52.26} & \cellcolor{gray!20}\textbf{54.59} & \cellcolor{gray!20}\textbf{63.02} & \cellcolor{gray!20}\textbf{65.20} & \cellcolor{gray!20}\textbf{48.19} & \cellcolor{gray!20}\textbf{65.09} & \cellcolor{gray!20}\textbf{68.92} & \cellcolor{gray!20}\textbf{66.81} & \cellcolor{gray!20}\textbf{67.71} & \cellcolor{gray!20}\textbf{65.54} & \cellcolor{gray!20}\textbf{76.25} & \cellcolor{gray!20}\textbf{72.34} & \cellcolor{gray!20}\textbf{52.72} & \cellcolor{gray!20}\textbf{59.55} & \cellcolor{gray!20}\textbf{54.98} & \cellcolor{gray!20}\textbf{62.21} \\
\midrule
\multicolumn{17}{l}{\textbf{Tiny-ImageNet-C}} \\
\cmidrule(r){1-17}
CLIP & 17.26 & 21.57 & 17.32 & 35.63 & 13.62 & 50.11 & 43.65 & 48.04 & 49.36 & 37.58 & 55.85 & 5.78 & 34.15 & 45.90 & 50.13 & 34.98 \\
TENT & 26.60 & 28.93 & 22.22 & 38.66 & 10.98 & 52.53 & 46.02 & 49.13 & 50.46 & 38.63 & 58.25 & 8.66 & 36.81 & 50.77 & 51.88 & 40.28 \\
% SAR & & & & & & & & & & & & & & & & \\
% VTE & & & & & & & & & & & & & & & & \\
TPT & 23.22 & 27.39 & 19.79 & 39.44 & 15.72 & 53.91 & \textbf{48.82} & 51.50 & \underline{53.89} & 40.43 & 61.07 & 7.48 & \underline{40.65} & \underline{52.28} & 52.38 & 41.07 \\
WATT & 27.98 & 31.47 & 25.38 & \underline{40.55} & \underline{17.44} & \underline{54.46} & 48.28 & \underline{52.60} & 53.77 & \underline{42.57} & 62.01 & 9.21 & 39.85 & 51.83 & \underline{54.19} & \underline{43.28} \\
CLIPArTT & \underline{29.27} & \underline{33.01} & \underline{27.93} & 39.99 & 17.09 & 53.63 & 47.39 & 51.97 & 51.49 & 42.35 & 60.69 & \underline{11.25} & 38.44 & 52.22 & 52.46 & 42.98 \\
BATCLIP & 23.98 & 27.34 & 23.06 & 35.40 & 14.42 & 48.27 & 42.72 & 41.82 & 46.54 & 36.29 & 54.78 & 6.90 & 36.84 & 47.59 & 49.89 & 35.72 \\
\textbf{Ours} & \cellcolor{gray!20}\textbf{38.14} & \cellcolor{gray!20}\textbf{41.02} & \cellcolor{gray!20}\textbf{35.48} & \cellcolor{gray!20}\textbf{46.03} & \cellcolor{gray!20}\textbf{27.93} & \cellcolor{gray!20}\textbf{57.78} & \cellcolor{gray!20}\textbf{52.67} & \cellcolor{gray!20}\textbf{56.85} & \cellcolor{gray!20}\textbf{57.56} & \cellcolor{gray!20}\textbf{50.49} & \cellcolor{gray!20}\textbf{65.28} & \cellcolor{gray!20}\textbf{21.57} & \cellcolor{gray!20}\textbf{47.86} & \cellcolor{gray!20}\textbf{58.23} & \cellcolor{gray!20}\textbf{56.82} & \cellcolor{gray!20}\textbf{50.36} \\
\midrule
\multicolumn{17}{l}{\textbf{ImageNet-C}} \\
\cmidrule(r){1-17}
CLIP & 19.26 & 21.62 & 17.88 & 27.20 & 20.26 & 31.78 & 29.22 & 43.16 & 38.08 & 43.38 & 61.52 & 29.50 & 20.76 & 44.22 & 32.94 & 32.05 \\
TENT & 20.48 & 22.70 & 15.86 & 28.00 & 21.74 & 33.40 & 30.44 & 43.96 & 38.58 & 43.92 & 61.88 & 32.22 & 22.22 & 45.84 & 35.10 & 33.09 \\
% SAR & & & & & & & & & & & & & & & & \\
% VTE & & & & & & & & & & & & & & & & \\
TPT & 19.00 & 26.00 & 17.00 & 28.00 & 16.00 & 30.00 & 30.00 & 42.00 & 31.00 & 41.00 & 59.00 & 32.00 & 15.00 & 43.00 & 31.00 & 30.67 \\
WATT & 23.32 & 26.18 & 6.06 & 32.26 & 27.66 & 38.54 & 34.96 & \textbf{47.14} & \underline{41.52} & 46.46 & \underline{63.32} & \underline{38.66} & 26.92 & \underline{50.10} & 41.38 & 36.30 \\
CLIPArTT & \underline{27.10} & \underline{29.00} & \textbf{29.20} & 21.80 & 21.30 & 31.00 & 28.40 & 39.60 & 38.50 & 41.50 & \textbf{64.60} & 18.40 & \textbf{30.70} & 43.30 & \textbf{48.00} & 34.13 \\
BATCLIP & 19.26 & 22.64 & 16.50 & \underline{33.02} & \underline{29.12} & \textbf{40.50} & \textbf{36.32} & \underline{47.08} & 40.86 & \underline{48.10} & 63.28 & 37.28 & 27.88 & \textbf{50.56} & \underline{43.64} & \underline{37.07} \\
\textbf{Ours} & \cellcolor{gray!20}\textbf{27.68} & \cellcolor{gray!20}\textbf{29.50} & \cellcolor{gray!20}\underline{28.74} & \cellcolor{gray!20}\textbf{34.52} & \cellcolor{gray!20}\textbf{30.58} & \cellcolor{gray!20}\underline{39.08} & \cellcolor{gray!20}\underline{35.50} & \cellcolor{gray!20}\underline{47.08} & \cellcolor{gray!20}\textbf{42.26} & \cellcolor{gray!20}\textbf{48.48} & \cellcolor{gray!20}59.60 & \cellcolor{gray!20}\textbf{39.84} & \cellcolor{gray!20}\underline{30.20} & \cellcolor{gray!20}48.42 & \cellcolor{gray!20}40.88 & \cellcolor{gray!20}\textbf{38.82} \\
\bottomrule
\end{tabular}
\end{table*}

\section{Comparison to OT-VP}

The very recent OT-VP \cite{zhang2024ot} has presented a solution integrating optimal transport for test-time adaptation. Nevertheless, it presents several fundamental differences with our work. First, OT-VP involves learning a universal visual prompt for the target domain, for which an optimal transport distance is optimized. Thus, optimal transport is used for a different task. Second, OT-VP is tailored to only visual models, not being capable of leveraging the valuable information found on the text modality. And third, the adaptation scenario strongly differs from our setting. In particular, OT-VP first fine-tunes the pre-trained model (pre-trained on ImageNet-1k) to a subdomain (e.g., PACS) with labeled data, to later adapt at test-time to the other subdomains (e.g., OfficeHome). While this strategy can be done in a single-source setting, OT-VP also evaluates the performance when multiple domains are used as the source (e.g., supervised adaptation on PACS), and only one left out for testing, referred to as Multi-Source in Table \ref{tab:performance_comparison}. In contrast, we follow the protocol of recent literature of CLIP test-time adaptation, where CLIP is directly exposed to the unsupervised test data points, without intermediate adaptation steps. To expand the extent of our empirical validation, we compare our approach to OT-VP \cite{zhang2024ot} whose results are reported in Table \ref{tab:performance_comparison}. To ensure a rigorous comparison, we use the same visual encoder as the backbone used in OT-VP and other prior models, i.e., ViT-B/16.

\begin{table}[t!]
\caption{Performance comparison to \cite{zhang2024ot} across PACS and OfficeHome datasets. % with average gain. 
Best results highlighted in bold, whereas \diff{value} indicates the difference wrt OT-VP in both single (\textit{first value}) and multi (\textit{second value}) source scenarios.}
\label{tab:performance_comparison}
\centering
\scriptsize
\begin{tabular}{lcc|c}
\toprule
\textbf{Method} & \textbf{PACS}  & \textbf{OfficeHome} & \textbf{Avg.} \\
\midrule
\multicolumn{4}{c}{\textbf{First Setting : Single Source}} \\
\midrule
OT-VP-B$_{\textit{WACV'25}}$ & 69.8 &  66.9 & 67.3 \\
OT-VP$_{\textit{WACV'25}}$ & 73.5 & 68.1 & 70.0 \\
\midrule
\multicolumn{4}{c}{\textbf{Second Setting : Multi Source }} \\
\midrule
OT-VP-B$_{\textit{WACV'25}}$ & 87.3 & 74.3 & 80.6 \\
OT-VP$_{\textit{WACV'25}}$ & 87.7 & 75.1 & 81.2 \\
\midrule
\multicolumn{4}{c}{\textbf{Our setting: \textit{No specific source}}} \\
\midrule
%\Ours{}& 96.89 & 80.15 & 83.37 & 86.80 \\
\Ours{}(\textit{Ours}) & \textbf{96.9}\diff{+23.4/+9.2} & \textbf{83.4}\diff{+15.3/+8.3} & \textbf{86.8}\diff{+16.8/+5.6} \\
\bottomrule
\end{tabular}
\end{table}

\section{Discussion on the Two Hyperparameters of WATT and Motivation for \Ours}
WATT \cite{osowiechi2024watt} introduces two key hyperparameters: $L$ (number of adaptation iterations for each text embedding) and $M$ (number of repetitions of the weight averaging process). While these hyperparameters play an essential role in improving WATT's adaptation capabilities, they introduce significant computational and scalability challenges, particularly as the number of classes grows. The iterative nature of $L$ requires multiple forward and backward passes for every template, leading to substantial runtime overhead, especially for datasets with many classes like Tiny-ImageNet. Additionally, higher $L$ values, while effective for complex corruptions, risk overfitting to noisy pseudo-labels, while lower 
$L$ values may lead to under-adaptation. Similarly, 
$M$ stabilizes adaptation through repeated weight averaging, but its effectiveness diminishes beyond a certain point, with larger values providing negligible performance improvements while exponentially increasing runtime. This dependence on repeated updates for each class and template becomes especially prohibitive for datasets with a large number of classes, as demonstrated in Figure \ref{fig:runtimes}, where runtime increases from CIFAR-10 (10 classes) to Tiny-ImageNet (200 classes).

These limitations motivate our proposed approach, \Ours, which simplifies and optimizes the adaptation process, achieving robust performance without the drawbacks associated with $L$ and $M$. \Ours{} eliminates these limitations by leveraging the Sinkhorn algorithm for single-pass optimization and precomputing averaged class embeddings, removing the need for iterative updates and weight averaging. This not only drastically reduces computational overhead but also ensures scalable, robust adaptation across diverse corruption levels and datasets, offering a practical alternative to WATT’s hyperparameter-dependent framework.

\section{Differences wrt PLOT \texorpdfstring{\cite{chenplot}}{}}

Note that despite the apparent similarities, our work presents fundamental differences w.r.t. Prompt Learning with Optimal Transport (PLOT) \cite{chenplot}, which prevents from adding the later in the empirical evaluation. \textbf{a) Target task.} PLOT tackles few-shot adaptation, requiring labeled data to adapt no novel tasks, whereas our method studies test-time adaptation, where labels or test samples are not available. Thus PLOT \textit{cannot be compared in the TTA setting.} \textbf{b) Adaptation.} Our approach updates the layer norm parameters of CLIP visual encoder, whereas PLOT optimizes the input text prompts (i.e., it falls into the Prompt Learning category). \textbf{c) Use of OT.} PLOT leverages OT between learnable prompts and a set of local visual features, whose transport plan is directly used as predicted probabilities. In contrast, we compute the transportation cost between the whole image features and fixed text embeddings, which is later used as a pseudo-supervision to distill knowledge during test-time adaptation.

%% file: main.bib
@String(IJCV = {Int. J. Comput. Vis.})

@String(CVPR= {IEEE Conf. Comput. Vis. Pattern Recog.})

@String(ICCV= {Int. Conf. Comput. Vis.})

@String(ECCV= {Eur. Conf. Comput. Vis.})

@String(ICLR = {Int. Conf. Learn. Represent.})

@String(IJCV  = {IJCV})

@String(CVPR  = {CVPR})

@String(ICCV  = {ICCV})

@String(ECCV  = {ECCV})

@String(ICLR  = {ICLR})

@inproceedings{asanoself,
  title={Self-labelling via simultaneous clustering and representation learning},
  author={Asano, YM and Rupprecht, C and Vedaldi, A},
  booktitle={International Conference on Learning Representations (ICLR)},
  year = 2020
}

@inproceedings{radford2021learning,
  title={Learning transferable visual models from natural language supervision},
  author={Radford, Alec and Kim, Jong Wook and Hallacy, Chris and Ramesh, Aditya and Goh, Gabriel and Agarwal, Sandhini and Sastry, Girish and Askell, Amanda and Mishkin, Pamela and Clark, Jack and others},
  booktitle={International Conference on Machine Learning (ICML)},
  pages={8748--8763},
  year={2021}
}

@inproceedings{WiSE,
  title={Robust fine-tuning of zero-shot models},
  author={Mitchell Wortsman and Gabriel Ilharco and Jong Wook Kim and Mike Li and Simon Kornblith and Rebecca Roelofs and Raphael Gontijo-Lopes and Hannaneh Hajishirzi and Ali Farhadi and Hongseok Namkoong and Ludwig Schmidt},
  booktitle={Proceedings of the IEEE/CVF Conference on Computer Vision and Pattern Recognition (CVPR)},
  pages={7959--7971},
  year={2022}
}

@inproceedings{osowiechi2024watt,
  title={W{ATT}: Weight Average Test-Time Adaption of {CLIP}},
  author={Osowiechi, David and Noori, Mehrdad and Hakim, Gustavo Adolfo Vargas and Yazdanpanah, Moslem and Bahri, Ali and Cheraghalikhani, Milad and Dastani, Sahar and Beizaee, Farzad and Ayed, Ismail Ben and Desrosiers, Christian},
  booktitle={Advances in Neural Information Processing Systems (NeurIPS)},
  year={2024}
}

@article{clipad,
   author = {Peng Gao and Shijie Geng and Renrui Zhang and Teli Ma and Rongyao Fang and Yongfeng Zhang and Hongsheng Li and Yu Qiao},
   journal = {International Journal of Computer Vision (IJCV)},
   title = {CLIP-Adapter: Better Vision-Language Models with Feature Adapters},
   year = {2023},
}

@inproceedings{jia2021scaling,
  title={Scaling up visual and vision-language representation learning with noisy text supervision},
  author={Jia, Chao and Yang, Yinfei and Xia, Ye and Chen, Yi-Ting and Parekh, Zarana and Pham, Hieu and Le, Quoc and Sung, Yun-Hsuan and Li, Zhen and Duerig, Tom},
  booktitle={International conference on machine learning (ICML)},
  pages={4904--4916},
  year={2021}
}

@inproceedings{silva2024closer,
  title={A closer look at the few-shot adaptation of large vision-language models},
  author={Silva-Rodriguez, Julio and Hajimiri, Sina and Ben Ayed, Ismail and Dolz, Jose},
  booktitle={Proceedings of the IEEE/CVF Conference on Computer Vision and Pattern Recognition (CVPR)},
  pages={23681--23690},
  year={2024}
}

@inproceedings{yu2023task,
  title={Task residual for tuning vision-language models},
  author={Yu, Tao and Lu, Zhihe and Jin, Xin and Chen, Zhibo and Wang, Xinchao},
  booktitle={Proceedings of the IEEE/CVF Conference on Computer Vision and Pattern Recognition (CVPR)},
  pages={10899--10909},
  year={2023}
}

@article{cuturi2013sinkhorn,
  title={Sinkhorn distances: Lightspeed computation of optimal transport},
  author={Cuturi, Marco},
  journal={Advances in neural information processing systems (NeurIPS)},
  volume={26},
  year={2013}
}

@article{knight2008sinkhorn,
  title={The Sinkhorn--Knopp algorithm: convergence and applications},
  author={Knight, Philip A},
  journal={SIAM Journal on Matrix Analysis and Applications},
  volume={30},
  number={1},
  pages={261--275},
  year={2008},
  publisher={SIAM}
}

@inproceedings{wangtent,
  title={Tent: Fully Test-Time Adaptation by Entropy Minimization},
  author={Wang, Dequan and Shelhamer, Evan and Liu, Shaoteng and Olshausen, Bruno and Darrell, Trevor},
  booktitle={International Conference on Learning Representations (ICLR)},
  year={2021}
}

@inproceedings{huang2019unsupervised,
  title={Unsupervised deep learning by neighbourhood discovery},
  author={Huang, Jiabo and Dong, Qi and Gong, Shaogang and Zhu, Xiatian},
  booktitle={International Conference on Machine Learning (ICML)},
  pages={2849--2858},
  year={2019}
}

@article{jabi2019deep,
  title={Deep clustering: On the link between discriminative models and k-means},
  author={Jabi, Mohammed and Pedersoli, Marco and Mitiche, Amar and Ayed, Ismail Ben},
  journal={IEEE transactions on pattern analysis and machine intelligence},
  volume={43},
  number={6},
  pages={1887--1896},
  year={2019},
  publisher={IEEE}
}

@inproceedings{niutowards,
  title={Towards Stable Test-time Adaptation in Dynamic Wild World},
  author={Niu, Shuaicheng and Wu, Jiaxiang and Zhang, Yifan and Wen, Zhiquan and Chen, Yaofo and Zhao, Peilin and Tan, Mingkui},
  booktitle={International Conference on Learning Representations (ICLR)},
  year={2022},
}

@inproceedings{choi2022improving,
  title={Improving test-time adaptation via shift-agnostic weight regularization and nearest source prototypes},
  author={Choi, Sungha and Yang, Seunghan and Choi, Seokeon and Yun, Sungrack},
  booktitle={European Conference on Computer Vision (ECCV)},
  pages={440--458},
  year={2022}
}

@article{zhang2024ot,
  title={O{T}-{VP}: Optimal Transport-guided Visual Prompting for Test-Time Adaptation},
  author={Zhang, Yunbei and Mehra, Akshay and Hamm, Jihun},
  journal={arXiv preprint arXiv:2407.09498},
  year={2024}
}

@article{zhang2022memo,
  title={Memo: Test time robustness via adaptation and augmentation},
  author={Zhang, Marvin and Levine, Sergey and Finn, Chelsea},
  journal={Advances in neural information processing systems (NeurIPS)},
  volume={35},
  pages={38629--38642},
  year={2022}
}

@inproceedings{yang2016joint,
  title={Joint unsupervised learning of deep representations and image clusters},
  author={Yang, Jianwei and Parikh, Devi and Batra, Dhruv},
  booktitle={Proceedings of the IEEE conference on computer vision and pattern recognition (CVPR)},
  pages={5147--5156},
  year={2016}
}

@inproceedings{yan2020clusterfit,
  title={Clusterfit: Improving generalization of visual representations},
  author={Yan, Xueting and Misra, Ishan and Gupta, Abhinav and Ghadiyaram, Deepti and Mahajan, Dhruv},
  booktitle={Proceedings of the IEEE/CVF Conference on Computer Vision and Pattern Recognition (CVPR)},
  pages={6509--6518},
  year={2020}
}

@inproceedings{caron2019unsupervised,
  title={Unsupervised pre-training of image features on non-curated data},
  author={Caron, Mathilde and Bojanowski, Piotr and Mairal, Julien and Joulin, Armand},
  booktitle={Proceedings of the IEEE/CVF International Conference on Computer Vision (ICCV)},
  pages={2959--2968},
  year={2019}
}

@inproceedings{caron2018deep,
  title={Deep clustering for unsupervised learning of visual features},
  author={Caron, Mathilde and Bojanowski, Piotr and Joulin, Armand and Douze, Matthijs},
  booktitle={European conference on computer vision (ECCV)},
  pages={132--149},
  year={2018}
}

@article{ma2024swapprompt,
  title={Swapprompt: Test-time prompt adaptation for vision-language models},
  author={Ma, Xiaosong and Zhang, Jie and Guo, Song and Xu, Wenchao},
  journal={Advances in Neural Information Processing Systems (NeurIPS)},
  volume={36},
  year={2023}
}

@inproceedings{dobler2024lost,
  title={A Lost Opportunity for Vision-Language Models: A Comparative Study of Online Test-time Adaptation for Vision-Language Models},
  author={D{\"o}bler, Mario and Marsden, Robert A and Raichle, Tobias and Yang, Bin},
   booktitle = {European Conference on Computer Vision (ECCV) Workshops},
  year={2024}
}

@inproceedings{yuan2023robust,
  title={Robust test-time adaptation in dynamic scenarios},
  author={Yuan, Longhui and Xie, Binhui and Li, Shuang},
  booktitle={Proceedings of the IEEE/CVF Conference on Computer Vision and Pattern Recognition (CVPR)},
  pages={15922--15932},
  year={2023}
}

@article{iwasawa2021test,
  title={Test-time classifier adjustment module for model-agnostic domain generalization},
  author={Iwasawa, Yusuke and Matsuo, Yutaka},
  journal={Advances in Neural Information Processing Systems},
  volume={34},
  pages={2427--2440},
  year={2021}
}

@inproceedings{lai2023padclip,
  title={Padclip: Pseudo-labeling with adaptive debiasing in clip for unsupervised domain adaptation},
  author={Lai, Zhengfeng and Vesdapunt, Noranart and Zhou, Ning and Wu, Jun and Huynh, Cong Phuoc and Li, Xuelu and Fu, Kah Kuen and Chuah, Chen-Nee},
  booktitle={Proceedings of the IEEE/CVF International Conference on Computer Vision (ICCV)},
  pages={16155--16165},
  year={2023}
}

@inproceedings{goyal2023finetune,
  title={Finetune like you pretrain: Improved finetuning of zero-shot vision models},
  author={Goyal, Sachin and Kumar, Ananya and Garg, Sankalp and Kolter, Zico and Raghunathan, Aditi},
  booktitle={Proceedings of the IEEE/CVF Conference on Computer Vision and Pattern Recognition (CVPR)},
  pages={19338--19347},
  year={2023}
}

@article{caron2020unsupervised,
  title={Unsupervised learning of visual features by contrasting cluster assignments},
  author={Caron, Mathilde and Misra, Ishan and Mairal, Julien and Goyal, Priya and Bojanowski, Piotr and Joulin, Armand},
  journal={Advances in neural information processing systems (NeurIPS)},
  volume={33},
  pages={9912--9924},
  year={2020}
}

@inproceedings{liang2020we,
  title={Do we really need to access the source data? source hypothesis transfer for unsupervised domain adaptation},
  author={Liang, Jian and Hu, Dapeng and Feng, Jiashi},
  booktitle={International conference on machine learning (ICML)},
  pages={6028--6039},
  year={2020}
}

@article{goyal2022test,
  title={Test time adaptation via conjugate pseudo-labels},
  author={Goyal, Sachin and Sun, Mingjie and Raghunathan, Aditi and Kolter, J Zico},
  journal={Advances in Neural Information Processing Systems (NeurIPS)},
  volume={35},
  pages={6204--6218},
  year={2022}
}

@inproceedings{hakim2024clipartt,
  title={C{LIPA}r{TT}: Light-weight Adaptation of {CLIP} to New Domains at Test Time},
  author={Hakim, Gustavo Adolfo Vargas and Osowiechi, David and Noori, Mehrdad and Cheraghalikhani, Milad and Bahri, Ali and Yazdanpanah, Moslem and Ayed, Ismail Ben and Desrosiers, Christian},
  booktitle={Proceedings of the IEEE/CVF Winter Conference on Applications of Computer Vision},
  year={2025}
}

@article{shu2022test,
  title={Test-time prompt tuning for zero-shot generalization in vision-language models},
  author={Shu, Manli and Nie, Weili and Huang, De-An and Yu, Zhiding and Goldstein, Tom and Anandkumar, Anima and Xiao, Chaowei},
  journal={Advances in Neural Information Processing Systems (NeurIPS)},
  volume={35},
  pages={14274--14289},
  year={2022}
}

@inproceedings{niu2022efficient,
  title={Efficient test-time model adaptation without forgetting},
  author={Niu, Shuaicheng and Wu, Jiaxiang and Zhang, Yifan and Chen, Yaofo and Zheng, Shijian and Zhao, Peilin and Tan, Mingkui},
  booktitle={International conference on machine learning (ICML)},
  pages={16888--16905},
  year={2022}
}

@inproceedings{mirza2022norm,
  title={The norm must go on: Dynamic unsupervised domain adaptation by normalization},
  author={Mirza, M Jehanzeb and Micorek, Jakub and Possegger, Horst and Bischof, Horst},
  booktitle={Proceedings of the IEEE/CVF conference on computer vision and pattern recognition (CVPR)},
  pages={14765--14775},
  year={2022}
}

@article{schneider2020improving,
  title={Improving robustness against common corruptions by covariate shift adaptation},
  author={Schneider, Steffen and Rusak, Evgenia and Eck, Luisa and Bringmann, Oliver and Brendel, Wieland and Bethge, Matthias},
  journal={Advances in neural information processing systems (NeurIPS)},
  volume={33},
  pages={11539--11551},
  year={2020}
}

@inproceedings{datasets_corruptions,
  author    = {Hendrycks, Dan and Dietterich, Thomas G.},
  title     = {Benchmarking Neural Network Robustness to Common Corruptions and Surface Noise},
  booktitle = {International Conference on Learning Representations (ICLR)},
  year      = {2019}
}

@inproceedings{cifar,
author = {Krizhevsky, Alex},
year = {2012},
title = {Learning Multiple Layers of Features from Tiny Images},
booktitle = {Preprint}
}

@inproceedings{cifar101,
  author       = {Benjamin Recht and
                  Rebecca Roelofs and
                  Ludwig Schmidt and
                  Vaishaal Shankar},
  title        = {Do {CIFAR-10} Classifiers Generalize to CIFAR-10?},
  booktitle      =  {Preprint},
  volume       = {abs/1806.00451},
  year         = {2018},
}

@inproceedings{TinyImagenet,
  title={Tiny ImageNet Challenge},
  author={Jiayu Wu and Qixiang Zhang and Guoxi Xu},
  year={2017},
  booktitle={Preprint},
  url={https://api.semanticscholar.org/CorpusID:212697711}
}

@inproceedings{pacs,
  title={Deeper, broader and artier domain generalization},
  author={Li, Da and Yang, Yongxin and Song, Yi-Zhe and Hospedales, Timothy M},
  booktitle={Proceedings of the IEEE International Conference on Computer Vision (ICCV)},
  pages={5542--5550},
  year={2017}
}

@inproceedings{vlcs,
  title={Unbiased metric learning: On the utilization of multiple datasets and web images for softening bias},
  author={Fang, Chen and Xu, Ye and Rockmore, Daniel N},
  booktitle={Proceedings of the IEEE International Conference on Computer Vision (ICCV)},
  pages={1657--1664},
  year={2013}
}

@inproceedings{officehome,
  title={Deep hashing network for unsupervised domain adaptation},
  author={Venkateswara, Hemanth and Eusebio, Jose and Chakraborty, Shayok and Panchanathan, Sethuraman},
  booktitle={Proceedings of the IEEE Conference on Computer Vision and Pattern Recognition (CVPR)},
  pages={5018--5027},
  year={2017}
}

@InProceedings{visda,
author = {Peng, Xingchao and Usman, Ben and Kaushik, Neela and Wang, Dequan and Hoffman, Judy and Saenko, Kate},
title = {VisDA: A Synthetic-to-Real Benchmark for Visual Domain Adaptation},
booktitle = {Proceedings of the IEEE Conference on Computer Vision and Pattern Recognition (CVPR) Workshops},
month = {June},
year = {2018}
}

@inproceedings{multimodal_dist_ot,
author = {Lee, John and Dabagia, Max and Dyer, Eva and Rozell, Christopher},
year = {2019},
booktitle={Advances in Neural Information Processing Systems (NeurIPS)},
title = {Hierarchical Optimal Transport for Multimodal Distribution Alignment},
doi = {10.48550/arXiv.1906.11768}
}

@inproceedings{chenplot,
  title={P{LOT}: Prompt Learning with Optimal Transport for Vision-Language Models},
  author={Chen, Guangyi and Yao, Weiran and Song, Xiangchen and Li, Xinyue and Rao, Yongming and Zhang, Kun},
  booktitle={The Eleventh International Conference on Learning Representations},
year={2023}
}

@misc{zhai2023sigmoidlosslanguageimage,
      title={Sigmoid Loss for Language Image Pre-Training}, 
      author={Xiaohua Zhai and Basil Mustafa and Alexander Kolesnikov and Lucas Beyer},
      year={2023},
      eprint={2303.15343},
      archivePrefix={arXiv},
      primaryClass={cs.CV},
      url={https://arxiv.org/abs/2303.15343}, 
}

@inproceedings{maharana2024texttt,
  title={B{ATCLIP}: Bimodal Online Test-Time Adaptation for CLIP},
  author={Maharana, Sarthak Kumar and Zhang, Baoming and Karlinsky, Leonid and Feris, Rogerio and Guo, Yunhui},
  booktitle={Proceedings of the IEEE/CVF International Conference on Computer Vision (ICCV)},
  year={2025}
}
